\documentclass{article}

\PassOptionsToPackage{numbers, compress}{natbib}
\usepackage[preprint]{neurips_2025}

\usepackage[utf8]{inputenc} 
\usepackage[T1]{fontenc}    
\usepackage{url}            
\usepackage{booktabs}       
\usepackage{amsfonts}       
\usepackage{nicefrac}       
\usepackage{microtype}      
\usepackage{xcolor}         
\usepackage[numbers]{natbib}
\usepackage{latexsym}
\usepackage{enumitem}
\usepackage[T1]{fontenc}

\usepackage[utf8]{inputenc}

\usepackage{microtype}

\usepackage{inconsolata}

\usepackage{graphicx}

\definecolor{cvprblue}{rgb}{0.21,0.49,0.74}

\usepackage{tcolorbox}
\tcbuselibrary{breakable}
\usepackage{booktabs}
\usepackage{graphicx}
\usepackage{multirow}
\usepackage{amsmath}
\usepackage{float}
\usepackage{makecell}
\usepackage{enumitem}
\usepackage{colortbl}
\usepackage{wrapfig}
\usepackage[pagebackref,breaklinks,colorlinks,allcolors=cvprblue]{hyperref}

\begin{document}
\title{MLLM-Guided VLM Fine-Tuning with Joint Inference for Zero-Shot Composed Image Retrieval}

%

\author{
  \textbf{Rong-Cheng Tu\textsuperscript{1}},
  \textbf{Zhao Jin\textsuperscript{1}},
  \textbf{Jingyi Liao\textsuperscript{1}},
  \textbf{Xiao Luo\textsuperscript{2}},
\\
  \textbf{Yingjie Wang\textsuperscript{1}},
  \textbf{Li Shen\textsuperscript{3}},
  \textbf{Dacheng Tao\textsuperscript{1}\footnotemark[1]}
\\
  \textsuperscript{1}College of Computing and Data Science, \\ Nanyang Technological University, Singapore, Singapore
\\
  \textsuperscript{2}Department of Computer Science, University of California, Los Angeles, USA
\\
  \textsuperscript{3}Sun Yat-sen University Shenzhen Campus, \\ School of Cyber Science and Technology, Shenzhen, China
\\
    \texttt{rongcheng.tu@gmail.com,} \texttt{dacheng.tao@ntu.edu.sg}
}

\renewcommand{\thefootnote}{\fnsymbol{footnote}}

\footnotetext[1]{\quad Co-corresponding authors}

\maketitle

\begin{abstract}
Zero-Shot Composed Image Retrieval (ZS-CIR) aims to retrieve a target image based on a compositional query consisting of a reference image and a natural language modification, without relying on labeled training triplets.
Existing ZS-CIR methods typically train adapters that convert reference images into pseudo-text tokens, which are concatenated with the modifying text and processed by frozen text encoders in pretrained VLMs or LLMs. While this design leverages the strengths of large pretrained models, it only supervises the adapter to produce encoder-compatible tokens that loosely preserve visual semantics. Crucially, it does not directly optimize the composed query representation to capture the full intent of the composition or to align with the target semantics, thereby limiting retrieval performance—particularly in cases involving fine-grained or complex visual transformations.
To address this problem, we propose MLLM-Guided VLM Fine-Tuning with Joint Inference (MVFT-JI), a novel approach that leverages a pretrained multimodal large language model (MLLM) to construct two complementary training tasks using only unlabeled images. First, a \emph{target text retrieval task} aligns composed query representations with generated target text representations. Second, a \emph{text-to-image retrieval task} enhances cross-modal consistency by aligning MLLM-generated captions with corresponding images.  By jointly optimizing these tasks, our method enables the VLM to inherently acquire robust compositional retrieval capabilities, supported by the provided theoretical justifications and empirical validation.
Furthermore, during inference, we further prompt the MLLM to generate target texts from composed queries and compute retrieval scores by integrating similarities between (i) the composed query and candidate images, and (ii) the MLLM-generated target text and candidate images. This strategy effectively combines the VLM’s semantic alignment strengths with the MLLM’s reasoning capabilities. Extensive experiments across three benchmark datasets demonstrate that MVFT-JI significantly outperforms existing ZS-CIR methods, \textit{e.g.}, achieving an 11\% absolute improvement in R@1 on the CIRR dataset compared to the state-of-the-art baseline MLLM-I2W.
\end{abstract}

\section{Introduction}
Image retrieval has traditionally relied on either content-based analysis \cite{crsi,mls3rduh,uhscm} or textual descriptions \cite{rcitr,ucmhmi,dcph} to locate desired images. While such methods are effective for simple queries, they struggle in scenarios requiring nuanced modifications to reference images—an increasingly common need in domains ranging from e-commerce to creative design. This limitation has led to the development of \emph{Composed Image Retrieval (CIR)}, which allows users to specify search intent by combining a reference image with a modifying text query. However, conventional CIR methods \cite{cjpa,cirr,circlt,li2023dual,xu2023multi,yang2023composed,limn} typically require extensive manually annotated triplets of the form <reference image, modifying text, target image>, which significantly hampers their deployment across diverse domains.

Zero-shot CIR (ZS-CIR) methods have emerged as a promising solution to reduce reliance on large labeled datasets. Early methods \cite{pic2word,searle,fti4cir,keds,pm} often learn an adapter on unlabeled images for pretrained vision-language models (VLMs) \cite{clip,li2023blip,albef,scl_vl,glscl}. By mapping reference images into pseudo-text tokens, these adapters convert image–text composed queries into textual input, allowing the VLM to perform cross-modal retrieval and thus achieve CIR. Recent advances in large language models (LLMs) \cite{minicpm_llm,llama} and multimodal LLMs (MLLMs) \cite{minicpm,interlm,wang2024mpo,internvl,spagent,llma_survey,m2ramg} have demonstrated powerful reasoning and instruction-following capabilities, spurring new ZS-CIR research that leverages these strengths. For instance, some methods \cite{mcl,fromage,instructcir} directly utilize LLMs to build <image, caption, modification text, target text> quadruplets from image-caption pairs and train LLM-specific adapters, transforming images into pseudo-text tokens for input to the LLM, thereby enabling it to handle CIR tasks.

However, existing approaches suffer from a fundamental limitation in their training paradigm. While the pseudo-text tokens generated by trained adapters are compatible with the text encoders of pretrained LLMs or VLMs, they fail to capture the rich semantic nuances of visual content. Furthermore, their training paradigm lacks direct supervision for aligning the composed query with its target feature. As a result, the final composed representations often fail to capture the intended semantic information accurately—especially in cases involving fine-grained or compositional changes—thereby limiting retrieval effectiveness.

In this paper, we propose MLLM-Guided VLM Fine-Tuning with Joint Inference (MVFT-JI) to address these challenges. Instead of solely relying on adapters and pseudo-text representations, our method introduces two complementary fine-tuning tasks directly applicable to pretrained VLMs, using only unlabeled images. Specifically, our theoretical analysis shows that jointly optimizing these two tasks naturally equips the VLM with robust compositional retrieval capabilities without explicit supervision.
First, we propose a \emph{Target Text Retrieval Task}. In this task, an MLLM is prompted to generate modification texts describing semantic transformations of reference images, and then further prompted to generate detailed textual descriptions of the modified (target) images. Training the VLM on these synthetically generated triplets enables the model to effectively extract compositional query features aligned with the generated textual descriptions. 
For the second task, \emph{Text-to-Image Retrieval}, we use the MLLM to generate detailed descriptive captions directly from unlabeled images and train the VLM to align these synthetic captions closely with their respective visual counterparts. Therefore, by jointly optimizing these two tasks, MVFT-JI learns rich, semantically meaningful representations for composed queries, while simultaneously refining alignment between synthesized textual descriptions and visual features.

During inference, we leverage the MLLM to generate target text descriptions from composed queries and refine retrieval by jointly considering (i) the similarity between the composed query and candidate images, and (ii) the alignment between the MLLM-generated target text and candidate images. This strategy effectively integrates the VLM’s strong cross-modal alignment with the MLLM’s advanced reasoning capabilities, significantly enhancing retrieval performance. 

To conclude, our contributions are summarized as follows:
\begin{itemize}[nosep]
\item We propose a novel ZS-CIR method MVFT-JI that directly fine-tunes the pretrained VLM through two tasks—Target Text Retrieval and Text-to-Image Retrieval—using supervision synthesized from an MLLM. We further provide a theoretical justification showing that, by jointly optimizing these tasks, MVFT-JI inherently acquires robust compositional retrieval capabilities.
\item We design an inference strategy that jointly integrates VLM-derived compositional features with MLLM-generated target descriptions to compute retrieval scores, effectively combining the alignment capacity of VLMs with the semantic reasoning ability of MLLMs.
\item Extensive experiments on three benchmark datasets demonstrate that our proposed MVFT-JI framework substantially outperforms state-of-the-art ZS-CIR methods, achieving superior retrieval accuracy across diverse domains.
\end{itemize}

\section{Related Work}
Traditional image retrieval techniques fall into two primary categories: content-based methods that compute visual similarity between images \cite{crsi,psldh,wglhh}, and text-based approaches that match textual queries to image captions or semantic embeddings \cite{rcitr,uchstm,daphcm}. While effective in isolation, these paradigms are inherently limited in scenarios where retrieval requires nuanced composition of visual and linguistic cues. To address this gap, \emph{Composed Image Retrieval (CIR)} \cite{zhang2023enhance, TianYuxin23Image,mcl, lincir, cirevl} has emerged as a new retrieval paradigm that enables users to search using a reference image and a modifying text query.
\subsection{Composed Image Retrieval (CIR)}
\vspace{-0.3cm}
To achieve this goal, the major of existing methods~\cite{tirg, tgcir, wen2024simple, zhang2023enhance, TianYuxin23Image, bai2023sentence}  need extensive annotated triplets $\langle\boldsymbol{x}_i, \boldsymbol{m}_i, \boldsymbol{x}_i^t\rangle$ to direct optimize a VLM to maximize the similarity between the representations of the composed query and the target image: \(\theta^* = \arg\max_{\theta} P(\boldsymbol{x}_i^t| \boldsymbol{x}_i, \boldsymbol{m}_i;\theta), \)
where $\theta$ denotes the set of parameters of the fine-tuned VLM. For example, TIRG~\cite{tirg} proposed a residual gating mechanism to fuse image and text representations. More recent works leverage powerful pretrained vision-language models (VLPs) to improve CIR performance. DQU-CIR~\cite{wen2024simple} designs a raw-data level multimodal fusion approach, shifting the multimodal fusion operation from feature level to raw-data level, which can fully leverage the capabilities of vision-language pre-trained models like CLIP to achieve promising CIR performance.
Despite these advancements, these CIR methods heavily rely on large-scale labeled triplets <\text{reference image}, \text{text}, \text{target image}>. 
Heavy dependence on manual annotations raises costs and restricts CIR models' generalization to diverse domains.

\vspace{-0.3cm}
\subsection{Zero-Shot Composed Image Retrieval (ZS-CIR)}
\vspace{-0.3cm}
\label{sec:zscir}
To address the reliance on labeled triplets, recent efforts have explored \emph{Zero-Shot Composed Image Retrieval (ZS-CIR)}, which aims to perform CIR without requiring annotated supervision. A common strategy in ZS-CIR is to train an adapter that projects reference images into pseudo-text tokens~\cite{keds, isearle, mllmi2w, contex12w, mcl, lincir, cirevl}, thereby transforming the multimodal composed query into a textual input.

Early approaches, such as Pic2Word~\cite{pic2word}, trained a lightweight visual adapter to map image features into pseudo-text representations, enabling retrieval within a frozen VLM. 
FTI4CIR~\cite{fti4cir} further improved upon this by generating multiple pseudo-text tokens per image to better capture fine-grained visual semantics.
Recently, the Large Language Models (LLMs)~\cite{minicpm_llm, llama} are adopted to enhance compositional reasoning in CIR. CIReVL~\cite{cirevl} reframes CIR as a text-to-image retrieval task by prompting an LLM to merge a reference image caption with modifying text, generating a single textual query. MLLM-I2W~\cite{mllmi2w} employs a multimodal LLM to refine pseudo-text tokens by enriching them with salient subject words, improving the textual representation of reference images. More recently, MCL~\cite{mcl} generates triplets in the form of <\text{reference image}, \text{text condition}, \text{target caption}> via a multimodal LLM and subsequently trains an adapter on large-scale datasets such as CC3M~\cite{cc3m}.

Despite these advancements, existing ZS-CIR methods rely on pseudo-text tokens, which fail to fully capture visual semantics and lack natural language properties, limiting retrieval performance. To address this, we propose MVFT-JI, which eliminates the need for pseudo-text mapping and directly fine-tunes a VLM using only unlabeled images. 
\section{Proposed Method\label{sec:task_formulation}}

\label{section:method_overview}
Given an unlabeled image dataset $\{\boldsymbol{x}_i\}_{i=1}^n$, our method aims to fine-tune a vision-language model (VLM) for zero-shot composed image retrieval (ZS-CIR) by jointly optimizing two complementary training tasks: \textit{target text retrieval} and \textit{text-to-image retrieval}. In this section, we first provide a theoretical justification demonstrating how jointly optimizing these two tasks allows the model to acquire compositional query capabilities without relying on explicitly labeled triplet data. Then, we describe an automatic procedure to generate suitable training samples directly from unlabeled images. Finally, we outline the training procedure and the inference protocol.

\subsection{Theoretical Justification of Proposed Training Tasks}

Formally, zero-shot composed image retrieval aims to retrieve the most semantically relevant target image $\boldsymbol{x}_i^t$ from a candidate set $\mathcal{C}$ given a composed query $( \boldsymbol{x}_i, \boldsymbol{m}_i )$ consisting of a reference image $\boldsymbol{x}_i$ and a textual modification $\boldsymbol{m}_i$: 
\begin{equation} 
\boldsymbol{x}_i^t = \arg\max_{\boldsymbol{x}_c\in\mathcal{C}} P(\boldsymbol{x}_c \mid \boldsymbol{x}_i, \boldsymbol{m}_i;\theta),
\end{equation}
where $\theta$ denotes the parameters of the VLM.

To effectively address ZS-CIR without explicit annotations, we introduce a latent semantic assumption: for every composed query $( \boldsymbol{x}_i, \boldsymbol{m}_i )$, there exists a latent textual description $\boldsymbol{t}_i$ explicitly characterizing the semantic intent behind the retrieval of target image $\boldsymbol{x}_i^t$. Applying the law of total probability based on this assumption yields:
\begin{equation} 
\label{eq:decomposition}
P(\boldsymbol{x}_i^t \mid \boldsymbol{x}_i, \boldsymbol{m}_i;\theta) = \sum_{\boldsymbol{t}_i} P(\boldsymbol{x}_i^t \mid \boldsymbol{x}_i, \boldsymbol{m}_i, \boldsymbol{t}_i;\theta) P(\boldsymbol{t}_i \mid \boldsymbol{x}_i, \boldsymbol{m}_i;\theta).
\end{equation}

Since the latent textual description $\boldsymbol{t}_i$ completely encapsulates the intended semantics of the composed query, we further assume conditional independence between the target image $\boldsymbol{x}_i^t$ and the composed query $(\boldsymbol{x}_i,\boldsymbol{m}_i)$ conditioned on $\boldsymbol{t}_i$:
\begin{equation} 
\label{eq:independence}
P(\boldsymbol{x}_i^t \mid \boldsymbol{x}_i, \boldsymbol{m}_i, \boldsymbol{t}_i;\theta) = P(\boldsymbol{x}_i^t \mid \boldsymbol{t}_i;\theta).
\end{equation}

Substituting Eq.~\eqref{eq:independence} into Eq.~\eqref{eq:decomposition}, we simplify the retrieval probability as:
\begin{equation} 
\label{eq:simplified}
P(\boldsymbol{x}_i^t \mid \boldsymbol{x}_i, \boldsymbol{m}_i;\theta) = \sum_{\boldsymbol{t}_i} P(\boldsymbol{x}_i^t \mid \boldsymbol{t}_i;\theta) P(\boldsymbol{t}_i \mid \boldsymbol{x}_i, \boldsymbol{m}_i;\theta).
\end{equation}

Inspecting Eq.~\eqref{eq:simplified}, we observe that maximizing the retrieval probability $P(\boldsymbol{x}_i^t \mid \boldsymbol{x}_i, \boldsymbol{m}_i;\theta)$ can be naturally achieved by independently optimizing the two conditional probabilities $P(\boldsymbol{x}_i^t \mid \boldsymbol{t}_i;\theta)$ and $P(\boldsymbol{t}_i \mid \boldsymbol{x}_i, \boldsymbol{m}_i;\theta)$. Furthermore, these two conditional probabilities naturally correspond to two well-defined and complementary training tasks commonly employed in vision-language modeling:
\begin{itemize}[nosep]
    \item Maximizing $P(\boldsymbol{x}_i^t \mid \boldsymbol{t}_i;\theta)$ corresponds exactly to a \textbf{text-to-image retrieval} task, guiding the VLM to accurately retrieve target images given their associated textual descriptions.
    \item Maximizing $P(\boldsymbol{t}_i \mid \boldsymbol{x}_i, \boldsymbol{m}_i;\theta)$ aligns precisely with a \textbf{target text retrieval} task, guiding the VLM to retrieve appropriate textual descriptions for a given composed query (image plus textual modification).
\end{itemize}

Thus, the theoretical derivation explicitly demonstrates that jointly optimizing these two retrieval tasks inherently equips the model with robust compositional query capabilities for ZS-CIR without the need for explicitly labeled triplet data.

\subsection{Training Data Curation}
\label{section:data_generation} 
To effectively optimize the two conditional probabilities defined in Eq.\eqref{eq:simplified}, we automatically generate suitable training data directly from the unlabeled image dataset $\{\boldsymbol{x}_i\}_{i=1}^n$ using an MLLM. Given the lack of labeled annotations, we independently construct training samples tailored specifically for the two distinct tasks: (1) target text retrieval, which approximates $P(\boldsymbol{t}_i \mid \boldsymbol{x}_i, \boldsymbol{m}_i;\theta)$, and (2) text-to-image retrieval, which approximates $P(\boldsymbol{x}_i^t \mid \boldsymbol{t}_i;\theta)$. Consequently, the latent textual intents $\boldsymbol{t}_i$ for each task are generated separately and thus differ between these two tasks. Moreover, explicitly enumerating all potential latent textual intents $\boldsymbol{t}$ in Eq.\eqref{eq:simplified} is computationally infeasible. Therefore, we utilize the MLLM to produce only single representative textual intent $\boldsymbol{t}_i$ for each individual training instance. We detail the data generation process for each task in the following subsections.

\paragraph{Target Text Retrieval Data.}
For this task, we need to construct triplets $(\boldsymbol{x}_i, \boldsymbol{m}_i, \boldsymbol{t}_i)$ for each image $\boldsymbol{x}_i$. Here, $\boldsymbol{x}_i$ is adopted as the reference image, $\boldsymbol{m}_i$ is a generated modification text describing an intended transformation, and $\boldsymbol{t}_i$ is the corresponding generated target text, which describes the content of the image after applying the modification.

To generate $\boldsymbol{m}_i$, we design a modification text generation prompt $P_m$ (shown in Appendix \ref{sec:prompt}) and input it along with the image $\boldsymbol{x}_i$ into the MLLM:
\begin{equation}
\begin{aligned}
\boldsymbol{m}_i &= MLLM(\boldsymbol{x}_i, P_m).
\end{aligned}
\label{eq:pm}
\end{equation}
With the modification text $\boldsymbol{m}_i$, we further design a target text generation prompt $P_{tt}$ (shown in Appendix \ref{sec:prompt}) to guide the MLLM in producing a textual description $\boldsymbol{t}_i$ of the modified image content:
\begin{equation}
\begin{aligned}
\boldsymbol{t}_i &= MLLM(\boldsymbol{x}_i, \boldsymbol{m}_i, P_{tt}).
\end{aligned}
\label{eq:ptt}
\end{equation}
This process ensures that $\boldsymbol{t}_i$ explicitly captures the modified semantics of $\boldsymbol{x}_i$ based on $\boldsymbol{m}_i$.
The triplet \((\boldsymbol{x}_i, \boldsymbol{m}_i, \boldsymbol{t}_i)\) forms the core data for the target text retrieval task. It ensures the VLM to fuse textual and visual information of the composed query $(\boldsymbol{x}_i, \boldsymbol{m}_i)$ into a joint representation, aligning it with the corresponding target text $\boldsymbol{t}_i$, i.e., to model $P(\boldsymbol{t}_i \mid \boldsymbol{x}_i, \boldsymbol{m}_i;\theta)$.

\paragraph{Text-to-Image Retrieval Data.}
To model $P(\boldsymbol{x}_i^t \mid \boldsymbol{t}_i;\theta)$, we additionally generate descriptive captions $\boldsymbol{c}_i$ for each image $\boldsymbol{x}_i$ using an image-captioning prompt $P_c$ (shown in Appendix \ref{sec:prompt}):
\begin{equation}
\boldsymbol{c}_i = MLLM(\boldsymbol{x}_i, P_c).
\label{eq:caption_prompt}
\end{equation}
The resulting pairs $(\boldsymbol{c}_i, \boldsymbol{x}_i)$ serve as training data for the text-to-image retrieval task, enabling the model to ground textual semantics in corresponding visual representations. 

Together, these two types of training data support the joint optimization of the two decomposed components in Eq.~\eqref{eq:simplified}, allowing the model to learn the full mapping from composed queries to target images without any labeled triplet supervision.

\begin{figure*}[tp]
    \centering
    \includegraphics[width=\linewidth]{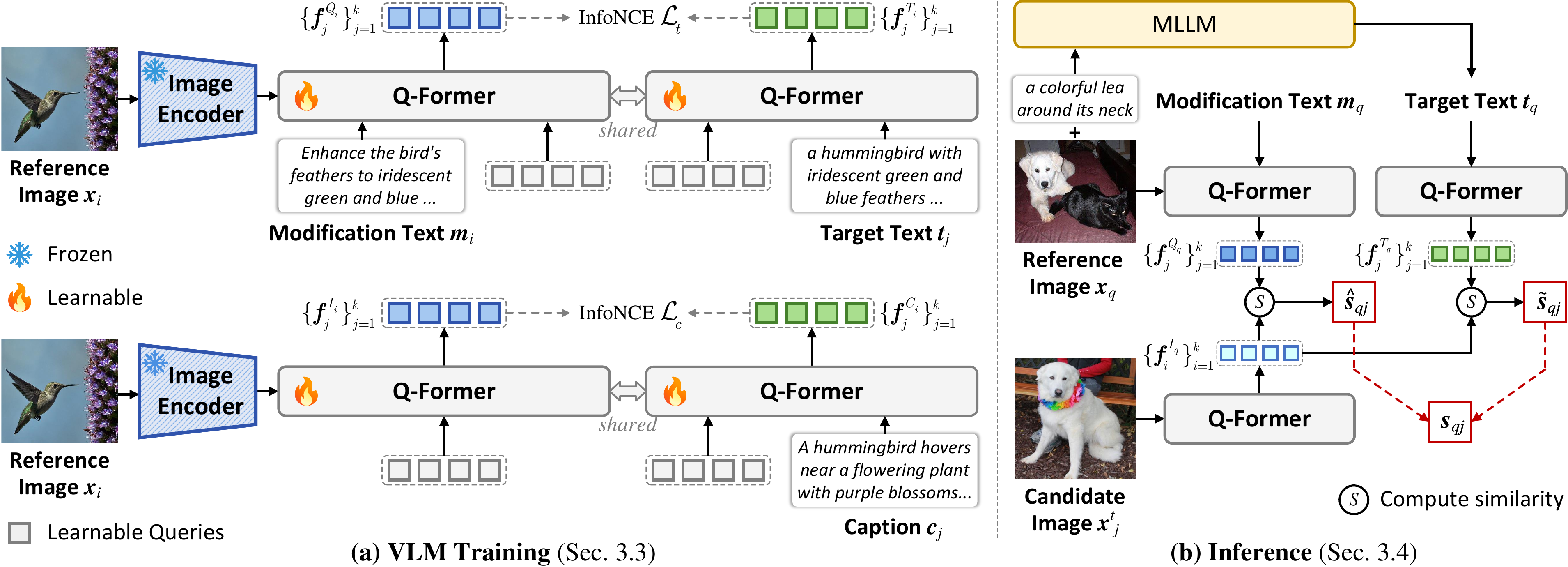}
    \vspace{-5pt}
    \caption{The training and inference framework of MVFT-JI.}
    \vspace{-5pt}
    \label{fig:training-inference}
\end{figure*}

\subsection{VLM Training}
\label{section:finetune_vlm}
With the constructed training data, we adopt a Q-Former-based vision-language model~\cite{li2023blip} and fine-tune it on the constructed data to jointly optimize the conditional probabilities in Eq.~\eqref{eq:simplified}.

\paragraph{Features extraction.} As shown in Figure \ref{fig:training-inference}, we employ a set of shared learnable query tokens \(\{\boldsymbol{q}_j\}_{j=1}^k\) to extract the representations of the composed queries, images and texts, respectively, where $k$ is the number of tokens. Specifically, for a composed query \(( \boldsymbol{x}_i, \boldsymbol{m}_i)\), we tokenize the modification text \(\boldsymbol{m}_i\) into text tokens, feed \(\boldsymbol{x}_i\) into a frozen image encoder to obtain image token features. Then we feed the visual and text tokens with the learnable query tokens \(\{\boldsymbol{q}_j\}_{j=1}^k\) into multiple Q-Former blocks, yielding the final feature tokens for the composed query \(( \boldsymbol{x}_i, \boldsymbol{m}_i)\):
\begin{equation}
    \{\boldsymbol{f}^{Q_i}_{j}\}_{j=1}^k = \mathrm{VLM}\big(\boldsymbol{x}_i, \boldsymbol{m}_i, \{\boldsymbol{q}_j\}_{j=1}^k\big).
\end{equation}
Similarly, for a target text \(\boldsymbol{t}_i\) and a image caption \(\boldsymbol{c}_i\), we feed only their text tokens with the learnable queries tokens \(\{\boldsymbol{q}_j\}_{j=1}^k\) into the Q-Former to generate their representation tokens, respectively: 
\begin{equation}
\begin{aligned}
    \{\boldsymbol{f}^{T_i}_{j}\}_{j=1}^k = \mathrm{VLM}\big(\boldsymbol{t}_i, \{\boldsymbol{q}_j\}_{j=1}^k\big), \ \ \ \ \{\boldsymbol{f}^{C_i}_{j}\}_{j=1}^k = \mathrm{VLM}\big(\boldsymbol{c}_i, \{\boldsymbol{q}_j\}_{j=1}^k\big),
\end{aligned}
\label{eq:similarity}
\end{equation}
For an image \(\boldsymbol{x}_i\), we first input it into a frozen image encoder to extract image token features, and then feed these features along with the query tokens into the Q-Former blocks to obtain the image's feature tokens:
\begin{equation}
\begin{aligned}
\{\boldsymbol{f}^{I_i}_{j}\}_{j=1}^k = \mathrm{VLM}\big(\boldsymbol{x}_i, \{\boldsymbol{q}_j\}_{j=1}^k\big).
\end{aligned}
\end{equation}

\paragraph{Similarity Computation.}
With the extracted features, we compute token-level cosine similarities and aggregate them by averaging the maximum token-pair similarities. Specifically, to measure the similarity $\hat{s}_{ij}$ between a composed query \(( \boldsymbol{x}_i, \boldsymbol{m}_i)\) and a generated target text \(\boldsymbol{t}_j\), we define:
\begin{equation}
    \hat{s}_{ij} = \frac{1}{k} \sum_{z=1}^k \max_{r} \frac{\big(\boldsymbol{f}^{Q_i}_{z}\big)^\top \boldsymbol{f}^{T_j}_{r}}{\|\boldsymbol{f}^{Q_i}_{z}\|_2 \|\boldsymbol{f}^{T_j}_{r}\|_2}, r\in\{1,\cdots , k\}.
\label{eq:sim_comp_query_text}
\end{equation}
Likewise, the similarity between a generated caption \(\boldsymbol{c}_i\) and an image \(\boldsymbol{x}_j\) is defined as follows:
\begin{equation}
    \tilde{s}_{ij} = \frac{1}{k} \sum_{z=1}^k \max_{r} \frac{\big(\boldsymbol{f}^{C_i}_{z}\big)^\top \boldsymbol{f}^{I_j}_{r}}{\|\boldsymbol{f}^{C_i}_{z}\|_2 \|\boldsymbol{f}^{I_j}_{r}\|_2}, r\in\{1,\cdots , k\}.
\label{eq:sim_image_caption}
\end{equation}

\paragraph{Softmax-based Probability Estimation.}
These similarity scores are transformed into normalized softmax distributions, enabling variational approximations of the conditional probabilities in Eq.~\eqref{eq:simplified}. Specifically, 1) The normalized similarity $\frac{\exp(\hat{s}_{ii}/\tau)}{\sum_j \exp(\hat{s}_{ij}/\tau)}$ estimates $P(\boldsymbol{t}_i \mid \boldsymbol{x}_i, \boldsymbol{m}_i;\theta)$; 2) The normalized similarity $\frac{\exp(\tilde{s}_{ii}/\tau)}{\sum_j \exp(\tilde{s}_{ij}/\tau)}$ estimates $P(\boldsymbol{x}_i^t \mid \boldsymbol{t}_i;\theta)$.
Thus, maximizing these probabilities via contrastive learning directly corresponds to maximizing the overall composed retrieval objective.

\paragraph{Training Objectives.}
The loss functions for the target text retrieval and text-to-image retrieval tasks are $\mathcal{L}_t$ and $\mathcal{L}_c$, respectively, which are  InfoNCE losses \cite{infonce} and defined as follows:
\begin{equation}
    \mathcal{L}_t = \frac{1}{N} \sum_{i=1}^N \log \frac{\exp(\hat{s}_{ii}/\tau)}{\sum_{j=1}^N \exp(\hat{s}_{ij}/\tau)}, \ \ \ \ \ \ \ \ \mathcal{L}_c = \frac{1}{N} \sum_{i=1}^N \log \frac{\exp(\tilde{s}_{ii}/\tau)}{\sum_{j=1}^N \exp(\tilde{s}_{ij}/\tau)}.
\end{equation}
The overall loss is the sum of both objectives:
\begin{equation}
    \mathcal{L} = \mathcal{L}_t + \mathcal{L}_c.
\end{equation}

By minimizing $\mathcal{L}$, we effectively maximize the two conditional components in Eq.~\eqref{eq:simplified}, thereby enabling the VLM to model $P(\boldsymbol{x}_i^t \mid \boldsymbol{x}_i, \boldsymbol{m}_i;\theta)$ without relying on any annotated triplet supervision.

\subsection{Inference}
\label{section:inference}
After training the VLM with the two retrieval objectives, we can apply it to perform the zero-shot composed image retrieval task.

\paragraph{Composed Query-to-Image Matching.}
Given a composed query $( \boldsymbol{x}_q, \boldsymbol{m}_q )$ consisting of a reference image and a textual modification, we first extract its feature tokens via the fine-tuned VLM:
\begin{equation}
\{\boldsymbol{f}^{Q_q}_i\}_{i=1}^k = \mathrm{VLM}(\boldsymbol{x}_q, \boldsymbol{m}_q,  \{\boldsymbol{q}_j\}_{j=1}^k).
\end{equation}
For each candidate image $\boldsymbol{x}^t_j \in \mathcal{C}$, we similarly obtain its image feature representation:
\begin{equation}
\{\boldsymbol{f}^{I_j}_i\}_{i=1}^k = \mathrm{VLM}(\boldsymbol{x}^t_j,  \{\boldsymbol{q}_j\}_{j=1}^k).
\end{equation}
We compute the similarity score $\hat{s}_{qj}$ between the composed query and each candidate image using the same token-level similarity in Eq.~\eqref{eq:sim_comp_query_text}.

Since the VLM has been trained to align composed queries with their target semantic information, $\hat{s}_{qj}$ provides a strong retrieval signal. The candidate with the highest similarity is expected to match the compositional intent of the query.

\paragraph{Semantic Reasoning via MLLM-Augmented Retrieval.}
To further enhance retrieval by leveraging the generative reasoning ability of MLLMs, we synthesize an imagined target description $\boldsymbol{t}_q$ for the composed query via the MLLM using the same prompt $P_{tt}$ employed during training:
\begin{equation}
\boldsymbol{t}_q = \mathrm{MLLM}(\boldsymbol{x}_q, \boldsymbol{m}_q, P_{tt}).
\label{inference_tt}
\end{equation}
This generated text serves as an additional semantic clue that captures the intended outcome of the composition. We encode $\boldsymbol{t}_q$ using the VLM:
\begin{equation}
\{\boldsymbol{f}^{T_q}_j\}_{j=1}^k = \mathrm{VLM}(\boldsymbol{t}_q, \{\boldsymbol{q}_j\}),
\end{equation}
and compute its similarity $\tilde{s}_{qj}$ to each candidate image $\boldsymbol{x}^t_j$ using Eq.~\eqref{eq:sim_image_caption}.

\paragraph{Fusion-based Retrieval Scoring.}
To combine the strengths of the fine-tuned VLM and the generative MLLM, we fuse the two similarity scores:
\begin{equation}
s_{qj} = \frac{1}{2}(\hat{s}_{qj} + \tilde{s}_{qj}).
\end{equation}
Finally, we rank all candidate images in $\mathcal{C}$ based on $s_{qj}$. This fusion strategy effectively integrates VLM-based compositional alignment with MLLM-driven semantic reasoning, yielding more robust and accurate retrieval performance in the CIR setting.

\section{Experiments}
\subsection{Experimental Setup  }

\begin{table*}[t]
  \centering 
  \vspace{-15pt}
  \caption{Results on Fashion-IQ validation set.}
  \label{tab:fashioniq}
  \resizebox{0.9\linewidth}{!}{
  \begin{tabular}{c|cc|cc|cc|cc} 
  \toprule
  \multirow{2}{*}{Methods} & \multicolumn{2}{c}{Shirt} & \multicolumn{2}{c}{Dress}& \multicolumn{2}{c}{TopTee} & \multicolumn{2}{c}{Average}\\
   \cmidrule(lr){2-3}
  \cmidrule(lr){4-5}
  \cmidrule(lr){6-7}
  \cmidrule(lr){8-9}
   & R@10 & R@50& R@10 & R@50& R@10 & R@50& R@10 & R@50\\  \midrule
  
Image-only  &10.40&22.03&3.91&12.14&7.70&18.05&7.33&17.41\\
  Text-only  &23.15&38.22&17.00&37.13&24.57&42.73&21.58&39.36\\
  Image$+$Text& 20.90&37.88&11.25&28.50&18.40&35.29&16.85&33.89\\
  Target Text& 21.39&39.49&15.66&36.04&21.67&42.78&19.57&39.44\\
  Pic2Word & 26.20 & 43.60 & 20.00 & 40.20 & 27.90 & 47.40 & 24.70 & 43.70 \\
  SEARLE-XL-OTI & {30.37} & 47.49 & 21.57 & 44.47 & 30.90 & 51.76 & 27.61 & 47.90 \\ 
  {iSEARLE-XL-OTI} & {31.80} & {50.20} & {24.19} & 45.12 & {31.72} & {53.29} & {29.24} & {49.54} \\
   {FTI4CIR} & 31.35 &50.59 & 24.39 & 47.84 & 32.43 & 54.21 & 29.39 & 50.88 \\ 
   Context-I2W & 29.70 & {48.60} & {23.10} & {45.30} & 30.60 & {52.90} & {27.80} & 48.93 \\
   CIReVL      & 29.49 & 47.40  & 24.79 & 44.76  & 31.36 & 53.65  & 28.55 & 48.57  \\ 
  LinCIR  & 29.10 & 46.81 & 20.92 & 42.44 & 28.81 & 50.18 & 26.28 & 46.49 \\
     MLLM-I2W  & 27.3 & 46.5 & 29.9 & 48.6 & 33.8 & 55.2 & 30.3 & 50.1 \\
    \rowcolor{gray!20}  MVFT-JI &\textbf{36.84} & \textbf{56.96} &\textbf{30.09} &\textbf{50.91} & \textbf{37.58} & \textbf{60.98} &\textbf{34.84} &\textbf{56.29} \\
\bottomrule
  \end{tabular}}
  \vspace{-10pt}
\end{table*}

\subsubsection{Evaluation Datasets and Protocol}
To evaluate the performance of our approach under diverse conditions, we adopt three public CIR benchmarks: \textbf{FashionIQ} \cite{fashioniq}, \textbf{CIRCO} \cite{searle}, and \textbf{CIRR} \cite{cirr}. The details of the three dataset are shown in the Appendix \ref{sec:prompt}.

We follow the standard evaluation metrics recommended for each dataset while highlighting their specific characteristics.
Consistent with prior studies \cite{fti4cir,lincir}, we report Recall at rank \(R@K\) (specifically \(K=10\) and \(50\)) for the FashionIQ dataset. We then average the results across the three fashion categories to assess overall performance.
Since each sample in CIRCO may have multiple valid target images, we follow \cite{fti4cir} and adopt the Mean Average Precision \(mAP@K\) metric (\(K = 5, 10, 25, 50\)). This choice offers a more fine-grained assessment of whether the system retrieves all relevant images.
For CIRR dataset, following \cite{fti4cir}, we employ multiple metrics, including \(R@K\) (\(K = 1, 5, 10, 50\)) and \(R_{\text{s}}@K\) (\(K = 1, 2, 3\)). The latter restricts the candidate targets to images that are semantically close to the correct target, thus alleviating the false-negative issue. We also compute the average of \(R@5\) and \(R_{\text{s}}@1\) as a single numerical measure of overall retrieval robustness.

\subsubsection{Baselines}
We compare our method against several ZS-CIR methods, including Pic2Word \cite{pic2word}, Context-I2W \cite{contex12w}, LinCIR \cite{lincir}, SEARLE \cite{searle}, iSEARLE-XL \cite{isearle}, FTI4CIR \cite{fti4cir}, CIReVL \cite{cirevl}, MLLM-I2W \cite{mllmi2w}, and MCL \cite{mcl}. Additionally, similar to previous methods \cite{mllmi2w,fti4cir}, we design the following baseline variants using BLIP2 encoders: 1) {Image-only}, 2) {Text-only}, 3) {Image + Text}, and 4) {Target Text}. The definitions of these variants are provided in the Appendix \ref{detial}.

\subsubsection{Implementation Details}
\label{sec:implementation}
Similar to existing CIR methods \cite{pic2word, fti4cir}, we use the subset of ImageNet1K \cite{imagenet}, contained  10K unlabeled images, as our fine-tuning dataset. We adopt \textbf{MiniCPM-V-2\_6} \cite{minicpm} as the MLLM to generate modifying texts, target texts, and image captions. Our Q-Former based vision-language model (VLM) is initialized with BLIP2 (ViT-L/14) \cite{li2023blip}.
For fine-tuning, we employ the AdamW optimizer \cite{adamw} with an initial learning rate of \(1 \times 10^{-5}\), which is decayed by a factor of 0.1 every 10 epochs. The model is trained with a batch size of 128. All experiments are conducted in PyTorch with fixed random seeds to ensure reproducibility. Notably, after pretraining on the unlabeled ImageNet1K subset, our VLM is evaluated on FashionIQ, CIRCO, and CIRR in a zero-shot setting—\textit{i.e.}, no additional dataset-specific fine-tuning is applied. For all experiments reported in the paper, we conducted the training and evaluation on a signle NVIDIA A100 40GB GPU. Each experiment was repeated three times with different random seeds, and the reported results are the average of these runs to ensure consistency.

\begin{table*}[ht]
\centering 
\vspace{-25pt}
\caption{Results on CIRR test set.}
\resizebox{0.9\linewidth}{!}{
\begin{tabular}{c|cccc|ccc|c}
\toprule
{Method} & {R@1} & {R@5} & {R@10} & {R@50} & {R\textsubscript{s}@1} & {R\textsubscript{s}@2} & {R\textsubscript{s}@3} & {Avg} \\ \midrule
Image-only &7.83& 24.51& 34.89&61.11& 20.99 & 41.30 & 60.84 &22.75 \\
  Text-only & 20.31 & 43.98 &55.61&78.43&60.46 &80.87&90.92 &52.22\\
  Image + Text & 10.55 & 32.53 & 45.47 &76.29 &29.93&53.86&72.48 &31.23\\
  Target Text & 20.00 & 46.15 &58.48&81.93 &57.40& 77.37 & 88.65 &51.78\\
  Pic2Word  & 23.90 & 51.70 & 65.00 & 87.80 & - & - & - & - \\ 
  SEARLE-XL-OTI & 24.87 & 52.31 & 66.29 & 88.58 & 53.80 & 74.31 & 86.94 & 53.06 \\ 
  iSEARLE-XL-OTI &  {25.40} & {54.05} & {67.47} & {88.92} & - & - & - & - \\ 
 LinCIR & 25.04  & 53.25 &  66.68 & -& 57.11  & 77.37  & 88.89 & -\\
  FTI4CIR & {25.90} & {55.61} & {67.66} & {89.66} & {55.21} & {75.88} & {87.98} & {55.41} \\ 
  CIReVL &24.55 &52.31 &64.92 &86.34 & 59.54 &79.88 &89.69 &- \\
  MCL &26.22 &56.84 & 70.00 &91.35 &61.45 &81.61&91.93&59.15 \\ 
  MLLM-I2W &28.3  &57.9  &70.2  &93.9  & - & - & - &-\\
  \rowcolor{gray!20} MVFT-JI & \textbf{39.30} & \textbf{69.49} &\textbf{79.69} &	\textbf{94.34}&\textbf{67.93 }&  \textbf{85.45} &   \textbf{93.13}&\textbf{68.71} \\ 
 \bottomrule
\end{tabular}}
\vspace{-13pt}
\label{tab:cirr}
\end{table*}

\subsection{Experimental results}
As shown in Table~\ref{tab:fashioniq}, we evaluate on FashionIQ dataset. Our approach significantly outperforms prior models on both Recall@10 and Recall@50, as well as on the average metric. For instance, compared to MLLM-I2W, we observe improvements of roughly 7--10\% in R@10 across all categories. 
Moreover, Table~\ref{tab:cirr} reports recall-based metrics (R@K and R\textsubscript{s}@K) on the CIRR test set. Our method attains the highest Recall@1 and R\textsubscript{s}@5, surpassing competitive baselines by up to 11\% and 6.48\% absolute gains, respectively. In particular, these improvements indicate that our model can more accurately retrieve the intended target image among closely related distractors. We attribute this strong performance primarily to the direct VLM fine-tuning with MLLM-generated triplets, which helps the VLM capture the target semantics from composed queries and align them with their corresponding target image features. Furthermore, the synergy between our fine-tuned VLM and MLLM-generated target text further boosts retrieval effectiveness.

To further assess ranking quality, Table~\ref{tab:circo} reports mean Average Precision (mAP) at various cutoffs on the CIRCO dataset. Once again, our method surpasses all baselines by a significant margin, achieving over a 3.12\% absolute improvement over the strongest competitor, CIReVL, at mAP@5. These superior mAP scores indicate that our retrieved results rank more relevant images higher compared to the baselines. This demonstrates that by capturing finer semantic nuances in both visual and textual representations, our model consistently produces a more precise ordering of candidates.

In conclusion, our method achieves the best retrieval performance on three established CIR benchmarks under zero-shot conditions. By unifying MLLM-based data generation with direct VLM fine-tuning and joint inference, we effectively bridge the semantic gap that has long hindered adapter-based CIR models. This performance boost demonstrates the promise of leveraging MLLMs not only for data augmentation but also for synergy in inference, paving the way for more robust and generalizable composed image retrieval systems in a wide range of domains.

\subsection{Ablation Study}
\vspace{-4pt}
To study the impact of each component in our \textit{MVFT-JI} framework, we design six variants: (1) \textbf{w/o $\tilde{s}_{qj}$} removes the MLLM-predicted target text based similarity, using only $\hat{s}_{qj}$ (i.e., the similarity between the fine-tuned VLM's composed-query features and candidate images) during inference; (2) \textbf{w/o $\hat{s}_{qj}$} discards the VLM-based similarity, relying solely on $\tilde{s}_{qj}$ (the similarity between MLLM-generated target text and candidate images); (3) \textbf{w/o $L_c$} omits the \emph{Text-to-Image Retrieval Task}, keeping only the \emph{Target Text Retrieval Task} for fine-tuning the VLM; (4) \textbf{w/o $L_t$} does the opposite, disabling the \emph{Target Text Retrieval Task} and retaining only the \emph{Text-to-Image Retrieval Task}; (5) \textbf{Internvl25} replaces our default MLLM (Minicpm-VL-2.6) with InternVL2\_5-8B \cite{internvl} when generating synthetic training data, to investigate the influence of different MLLM backbones; (6) \textbf{$\boldsymbol{P}'$} leverages a newly designed prompt set (see Appendix \ref{sec:prompt}) for MLLM data generation to examine how prompt engineering affects the final retrieval performance.
The results are summarized in Table~\ref{tab:ablation}.

\begin{wraptable}{r}{0.59\textwidth}
	\vspace{-25pt}
	\caption{Results on CIRCO test set.}
	\label{tab:circo}
	\vspace{4pt}
	\resizebox{\linewidth}{!}{%
		\begin{tabular}{c|cccc}
			\toprule
			{Method} & {mAP@5} & {mAP@10} & {mAP@25} & {mAP@50} \\ \hline
			Image-only &2.59&3.2&3.98& 4.52 \\
			Text-only &3.36&3.79&4.4&4.76 \\
			Image + Text &6.67 &7.98 &9.69&10.56 \\
			Captioning &10.27 &11.09 &12.53 &13.35 \\
			Pic2Word   & 8.72 & 9.51 & 10.46 & 11.29 \\
			SEARLE-XL-OTI  & 10.18 & 11.03 & 12.72 & 13.67 \\
			{iSEARLE-XL-OTI} & {11.31} & {12.67} & {14.46} & {15.34} \\
			LinCIR & 12.59 &13.58 & 15.00 &15.85 \\
			CIREVL & 18.57 & 19.01 & 20.89 & 21.80 \\
			MCL &17.67 &18.86 &20.80 & 21.68 \\
			FTI4CIR & {15.05} & {16.32} & {18.06} & {19.05} \\  
			\rowcolor{gray!20}  MVFT-JI &\textbf{21.69}&  \textbf{22.99}&  \textbf{25.35}&  \textbf{26.37} \\
			\bottomrule
		\end{tabular}
	}
	\vspace{-4pt}
\end{wraptable} 

\vspace{-5pt}
\paragraph{Joint Similarity vs. Single Similarity.}
Comparing MVFT-JI with \textbf{w/o $\tilde{s}_{qj}$} and \textbf{w/o $\hat{s}_{qj}$} clarifies the importance of combining both compositional features based similarity (\(\hat{s}_{qj}\)) and MLLM-generated text based similarity (\(\tilde{s}_{qj}\)) at inference. On FashionIQ, for instance, removing \(\tilde{s}_{qj}\) reduces R@10 from 34.84 to 32.52 (\textbf{w/o $\tilde{s}_{qj}$}), while ignoring $\hat{s}_{qj}$ plunges the score to 26.66 (\textbf{w/o $\hat{s}_{qj}$}). Similar declines appear for CIRO and CIRR. This indicates that joint inference is pivotal for robust retrieval, as it enriches the search process with both VLM's semantic alignment and MLLM's semantic reasoning capabilities.

\paragraph{Training Objective functions.}
We next examine the impact of removing each training objective on performance. When the Target Text Retrieval Task is excluded (\textbf{w/o $L_t$}), the model consistently performs worse across all benchmarks; for example, mAP@5 on CIRCO drops from 21.69 to 15.18. These results highlight the crucial role of fine-tuning the VLM to enhance its ability to extract the target semantic features of composed queries, which is essential for improving retrieval performance.
Similarly, removing the Text-to-Image Retrieval Task (\textbf{w/o $L_c$}) also leads to performance degradation on most benchmarks, with R@1 on CIRR decreasing from 39.30 to 36.15. However, the improvement brought by $L_c$ is less pronounced compared to $L_t$. We attribute this to the fact that the pretrained VLM already possesses strong cross-modal retrieval capabilities, thus additional fine-tuning with $L_c$ brings relatively smaller gains than $L_t$, which directly enhances composed query understanding. Interestingly, however, \textbf{w/o $L_c$} slightly outperforms the full model on CIRCO (mAP@5: 22.55 vs.\ 21.69). 
A possible explanation is that CIRCO allows multiple valid target images for a given composed query, whereas our training procedure generates one-to-one text-image pairs, introducing a distributional gap that may hinder performance when emphasizing strict text-to-image alignment. In contrast, while $L_t$ also relies on a single target text per query, its primary objective is to guide the model in extracting and representing the intended target semantics from composed queries, leading to more substantial overall improvements. As a result, even in CIRCO, models trained with $L_t$ consistently achieve better retrieval performance.

\begin{wraptable}{r}{0.7\textwidth}
\vspace{-15pt}
\caption{Ablation study on the three datasets. We report the average results of FashionIQ's  three categories.}
\label{tab:ablation}
\vspace{5pt}
\resizebox{\linewidth}{!}{%
\begin{tabular}{c|cc|cc|c|cc} 
  \toprule
  \multirow{2}{*}{Methods} & \multicolumn{2}{c}{FashionIQ} & \multicolumn{2}{c}{CIRO}& \multicolumn{2}{c}{CIRR} \\
   \cmidrule(lr){2-3}
  \cmidrule(lr){4-5}
  \cmidrule(lr){6-7}
   & R@10 & R@50& mAP@5 & mAP@10& R@1 & R@5\\  \midrule
  w/o $\tilde{s}_{qj}$  &32.52 &53.50 &17.53 &18.96 & 30.34 &60.10\\
  w/o $\hat{s}_{qj}$  &26.66&47.44&15.21&15.93  &33.11   &61.81\\
  w/o $L_c$&32.75&54.74&\textbf{22.55} &  \textbf{24.27}&36.15 & 66.36\\
  w/o $L_t$ &29.83&48.28& 15.18  & 16.28 &34.31 &65.23\\
  Internvl25 &31.88&54.16&18.18  &18.84&37.47  & 67.59\\
  $P'$ &33.72&55.69&22.15 &23.98&39.11&\textbf{69.81}\\
\rowcolor{gray!20}   MVFT-JI &\textbf{34.84} &\textbf{56.29}&21.69&  22.99& \textbf{39.30} & 69.49\\
\bottomrule
  \end{tabular}
}
\vspace{-10pt}
\end{wraptable} 
\paragraph{Replacing the MLLM Backbone.}
We use \textbf{Internvl25} to test how different multimodal language models affect data generation. Substituting InterVL-VL-2.5 for Minicpm-VL-2.6 leads to consistently lower scores on FashionIQ and CIRO (e.g., mAP@5 on CIRO drops from 21.69 to 18.18), suggesting that a more capable MLLM can synthesize more accurate textual training data. Consequently, the VLM benefits from higher-quality supervision to improve its retrieval performance.

\paragraph{Prompt Engineering Effects.}
In \textbf{$\boldsymbol{P}'$}, we replace the default prompt set with a new set of instructions (detailed in the Appendix \ref{sec:prompt}). This change yields  some improvement on CIRO (e.g., mAP@10 increases from 22.99 to 23.98). However, the results are not universally better across all datasets, pointing to the nuanced influence of prompt design on data style and diversity. Fine-tuning these prompts could further refine the balance between semantic specificity and generalizability.

Overall, these ablation results underscore three central points. First, combining VLM-based compositional similarity with MLLM-generated textual similarity is essential for capturing both visual and linguistic nuances. Second, our dual training objectives ($L_c$ and $L_t$) are broadly complementary, although some dataset-specific patterns (e.g., CIRO) may favor one objective slightly more. Finally, the choice of MLLM backbone and the design of prompts both significantly shape the quality of the synthetic data used for fine-tuning, which in turn affects zero-shot retrieval performance. 
\vspace{-0.2cm}
\section{Conclusion}
\label{sec:conclusion}
\vspace{-0.2cm}
In this paper, we introduced a novel ZS-CIR method \emph{MVFT-JI}. Unlike prior methods that rely on adapter-generated pseudo-text tokens, our approach leverages the descriptive and generative strengths of MLLMs to synthesize training data from unlabeled images for two complementary tasks: target text retrieval and text-to-image retrieval. We then directly fine-tune a VLM on these tasks to enhance its ability to learn semantically rich compositional representations, thereby enabling effective composed image retrieval. This training paradigm is supported by our provided theoretical justifications and empirical validation. At inference time, we further combine the VLM-derived compositional features with MLLM-generated target texts to compute similarity, ensuring robust alignment between composed queries and target images. Extensive experiments on three standard ZS-CIR benchmarks demonstrate that MVFT-JI consistently outperforms the state-of-the-art methods.

\bibliographystyle{unsrtnat}
\bibliography{neurips_2025}

\begin{thebibliography}{57}
\providecommand{\natexlab}[1]{#1}
\providecommand{\url}[1]{\texttt{#1}}
\expandafter\ifx\csname urlstyle\endcsname\relax
  \providecommand{\doi}[1]{doi: #1}\else
  \providecommand{\doi}{doi: \begingroup \urlstyle{rm}\Url}\fi

\bibitem[Chua et~al.(1994)Chua, Lim, and Pung]{crsi}
Tat{-}Seng Chua, S.{-}K. Lim, and Hung~Keng Pung.
\newblock Content-based retrieval of segmented images.
\newblock In Meera Blattner and John~O. Limb, editors, \emph{Proceedings of the
  Second {ACM} International Conference on Multimedia '94, San Francisco, CA,
  USA, October 15-20, 1994}, pages 211--218. {ACM} Press, 1994.
\newblock \doi{10.1145/192593.192658}.
\newblock URL \url{https://doi.org/10.1145/192593.192658}.

\bibitem[Tu et~al.(2020)Tu, Mao, and Wei]{mls3rduh}
Rong{-}Cheng Tu, Xianling Mao, and Wei Wei.
\newblock {MLS3RDUH:} deep unsupervised hashing via manifold based local
  semantic similarity structure reconstructing.
\newblock In Christian Bessiere, editor, \emph{Proceedings of the Twenty-Ninth
  International Joint Conference on Artificial Intelligence, {IJCAI} 2020},
  pages 3466--3472. ijcai.org, 2020.
\newblock \doi{10.24963/IJCAI.2020/479}.
\newblock URL \url{https://doi.org/10.24963/ijcai.2020/479}.

\bibitem[Tu et~al.(2023{\natexlab{a}})Tu, Mao, Lin, Cai, Qin, Wei, Wang, and
  Huang]{uhscm}
Rong{-}Cheng Tu, Xian{-}Ling Mao, Kevin~Qinghong Lin, Chengfei Cai, Weize Qin,
  Wei Wei, Hongfa Wang, and Heyan Huang.
\newblock Unsupervised hashing with semantic concept mining.
\newblock \emph{Proc. {ACM} Manag. Data}, 1\penalty0 (1):\penalty0 3:1--3:19,
  2023{\natexlab{a}}.
\newblock \doi{10.1145/3588683}.
\newblock URL \url{https://doi.org/10.1145/3588683}.

\bibitem[Rao et~al.(2022)Rao, Wang, Ding, Qi, Zhan, Liu, and Tao]{rcitr}
Jun Rao, Fei Wang, Liang Ding, Shuhan Qi, Yibing Zhan, Weifeng Liu, and Dacheng
  Tao.
\newblock Where does the performance improvement come from?: - {A}
  reproducibility concern about image-text retrieval.
\newblock In Enrique Amig{\'{o}}, Pablo Castells, Julio Gonzalo, Ben
  Carterette, J.~Shane Culpepper, and Gabriella Kazai, editors, \emph{{SIGIR}
  '22: The 45th International {ACM} {SIGIR} Conference on Research and
  Development in Information Retrieval, Madrid, Spain, July 11 - 15, 2022},
  pages 2727--2737. {ACM}, 2022.
\newblock \doi{10.1145/3477495.3531715}.
\newblock URL \url{https://doi.org/10.1145/3477495.3531715}.

\bibitem[Tu et~al.(2023{\natexlab{b}})Tu, Jiang, Lin, Cai, Tian, Wang, and
  Liu]{ucmhmi}
Rong{-}Cheng Tu, Jie Jiang, Qinghong Lin, Chengfei Cai, Shangxuan Tian, Hongfa
  Wang, and Wei Liu.
\newblock Unsupervised cross-modal hashing with modality-interaction.
\newblock \emph{{IEEE} Trans. Circuits Syst. Video Technol.}, 33\penalty0
  (9):\penalty0 5296--5308, 2023{\natexlab{b}}.
\newblock \doi{10.1109/TCSVT.2023.3251395}.
\newblock URL \url{https://doi.org/10.1109/TCSVT.2023.3251395}.

\bibitem[Tu et~al.(2023{\natexlab{c}})Tu, Mao, Tu, Bian, Cai, Wang, Wei, and
  Huang]{dcph}
Rong{-}Cheng Tu, Xian{-}Ling Mao, Rongxin Tu, Bin{-}Bin Bian, Chengfei Cai,
  Hongfa Wang, Wei Wei, and Heyan Huang.
\newblock Deep cross-modal proxy hashing.
\newblock \emph{{IEEE} Trans. Knowl. Data Eng.}, 35\penalty0 (7):\penalty0
  6798--6810, 2023{\natexlab{c}}.
\newblock \doi{10.1109/TKDE.2022.3187023}.
\newblock URL \url{https://doi.org/10.1109/TKDE.2022.3187023}.

\bibitem[Yang et~al.(2021)Yang, Wang, Zhou, and Li]{cjpa}
Yuchen Yang, Min Wang, Wengang Zhou, and Houqiang Li.
\newblock Cross-modal joint prediction and alignment for composed query image
  retrieval.
\newblock In Heng~Tao Shen, Yueting Zhuang, John~R. Smith, Yang Yang, Pablo
  C{\'{e}}sar, Florian Metze, and Balakrishnan Prabhakaran, editors, \emph{{MM}
  '21: {ACM} Multimedia Conference, Virtual Event, China, October 20 - 24,
  2021}, pages 3303--3311. {ACM}, 2021.
\newblock \doi{10.1145/3474085.3475483}.
\newblock URL \url{https://doi.org/10.1145/3474085.3475483}.

\bibitem[Liu et~al.(2021)Liu, Opazo, Teney, and Gould]{cirr}
Zheyuan Liu, Cristian~Rodriguez Opazo, Damien Teney, and Stephen Gould.
\newblock Image retrieval on real-life images with pre-trained
  vision-and-language models.
\newblock In \emph{2021 {IEEE/CVF} International Conference on Computer Vision,
  {ICCV} 2021, Montreal, QC, Canada, October 10-17, 2021}, pages 2105--2114.
  {IEEE}, 2021.
\newblock \doi{10.1109/ICCV48922.2021.00213}.
\newblock URL \url{https://doi.org/10.1109/ICCV48922.2021.00213}.

\bibitem[Baldrati et~al.(2024)Baldrati, Bertini, Uricchio, and Bimbo]{circlt}
Alberto Baldrati, Marco Bertini, Tiberio Uricchio, and Alberto~Del Bimbo.
\newblock Composed image retrieval using contrastive learning and task-oriented
  clip-based features.
\newblock \emph{{ACM} Trans. Multim. Comput. Commun. Appl.}, 20\penalty0
  (3):\penalty0 62:1--62:24, 2024.
\newblock \doi{10.1145/3617597}.
\newblock URL \url{https://doi.org/10.1145/3617597}.

\bibitem[Li(2023)]{li2023dual}
Shenshen Li.
\newblock Dual-path semantic construction network for composed query-based
  image retrieval.
\newblock In Ioannis Kompatsiaris, Jiebo Luo, Nicu Sebe, Angela Yao, Vasileios
  Mazaris, Symeon Papadopoulos, Adrian Popescu, and Zi~Helen Huang, editors,
  \emph{Proceedings of the 2023 {ACM} International Conference on Multimedia
  Retrieval, {ICMR} 2023, Thessaloniki, Greece, June 12-15, 2023}, pages
  636--639. {ACM}, 2023.
\newblock \doi{10.1145/3591106.3592245}.
\newblock URL \url{https://doi.org/10.1145/3591106.3592245}.

\bibitem[Xu et~al.(2023)Xu, Bin, Wei, Yang, Wang, and Shen]{xu2023multi}
Yahui Xu, Yi~Bin, Jiwei Wei, Yang Yang, Guoqing Wang, and Heng~Tao Shen.
\newblock Multi-modal transformer with global-local alignment for composed
  query image retrieval.
\newblock \emph{{IEEE} Trans. Multim.}, 25:\penalty0 8346--8357, 2023.
\newblock \doi{10.1109/TMM.2023.3235495}.
\newblock URL \url{https://doi.org/10.1109/TMM.2023.3235495}.

\bibitem[Yang et~al.(2023)Yang, Ye, Cai, Su, and Du]{yang2023composed}
Qu~Yang, Mang Ye, Zhaohui Cai, Kehua Su, and Bo~Du.
\newblock Composed image retrieval via cross relation network with hierarchical
  aggregation transformer.
\newblock \emph{{IEEE} Trans. Image Process.}, 32:\penalty0 4543--4554, 2023.
\newblock \doi{10.1109/TIP.2023.3299791}.
\newblock URL \url{https://doi.org/10.1109/TIP.2023.3299791}.

\bibitem[Wen et~al.(2024{\natexlab{a}})Wen, Song, Yin, Wu, Guan, and Nie]{limn}
Haokun Wen, Xuemeng Song, Jianhua Yin, Jianlong Wu, Weili Guan, and Liqiang
  Nie.
\newblock Self-training boosted multi-factor matching network for composed
  image retrieval.
\newblock \emph{{IEEE} Trans. Pattern Anal. Mach. Intell.}, 46\penalty0
  (5):\penalty0 3665--3678, 2024{\natexlab{a}}.

\bibitem[Saito et~al.(2023)Saito, Sohn, Zhang, Li, Lee, Saenko, and
  Pfister]{pic2word}
Kuniaki Saito, Kihyuk Sohn, Xiang Zhang, Chun{-}Liang Li, Chen{-}Yu Lee, Kate
  Saenko, and Tomas Pfister.
\newblock Pic2word: Mapping pictures to words for zero-shot composed image
  retrieval.
\newblock In \emph{{IEEE/CVF} Conference on Computer Vision and Pattern
  Recognition, {CVPR} 2023, Vancouver, BC, Canada, June 17-24, 2023}, pages
  19305--19314. {IEEE}, 2023.
\newblock \doi{10.1109/CVPR52729.2023.01850}.
\newblock URL \url{https://doi.org/10.1109/CVPR52729.2023.01850}.

\bibitem[Baldrati et~al.(2023)Baldrati, Agnolucci, Bertini, and Bimbo]{searle}
Alberto Baldrati, Lorenzo Agnolucci, Marco Bertini, and Alberto~Del Bimbo.
\newblock Zero-shot composed image retrieval with textual inversion.
\newblock In \emph{{IEEE/CVF} International Conference on Computer Vision,
  {ICCV} 2023, Paris, France, October 1-6, 2023}, pages 15292--15301. {IEEE},
  2023.
\newblock \doi{10.1109/ICCV51070.2023.01407}.
\newblock URL \url{https://doi.org/10.1109/ICCV51070.2023.01407}.

\bibitem[Lin et~al.(2024)Lin, Wen, Song, Liu, Hu, and Nie]{fti4cir}
Haoqiang Lin, Haokun Wen, Xuemeng Song, Meng Liu, Yupeng Hu, and Liqiang Nie.
\newblock Fine-grained textual inversion network for zero-shot composed image
  retrieval.
\newblock In Grace~Hui Yang, Hongning Wang, Sam Han, Claudia Hauff, Guido
  Zuccon, and Yi~Zhang, editors, \emph{Proceedings of the 47th International
  {ACM} {SIGIR} Conference on Research and Development in Information
  Retrieval, {SIGIR} 2024, Washington DC, USA, July 14-18, 2024}, pages
  240--250. {ACM}, 2024.
\newblock \doi{10.1145/3626772.3657831}.
\newblock URL \url{https://doi.org/10.1145/3626772.3657831}.

\bibitem[Suo et~al.(2024)Suo, Ma, Zhu, and Yang]{keds}
Yucheng Suo, Fan Ma, Linchao Zhu, and Yi~Yang.
\newblock Knowledge-enhanced dual-stream zero-shot composed image retrieval.
\newblock In \emph{{IEEE/CVF} Conference on Computer Vision and Pattern
  Recognition, {CVPR} 2024, Seattle, WA, USA, June 16-22, 2024}, pages
  26941--26952. {IEEE}, 2024.
\newblock \doi{10.1109/CVPR52733.2024.02545}.
\newblock URL \url{https://doi.org/10.1109/CVPR52733.2024.02545}.

\bibitem[Zhang et~al.(2024{\natexlab{a}})Zhang, Yanagi, Togo, Ogawa, and
  Haseyama]{pm}
Huaying Zhang, Rintaro Yanagi, Ren Togo, Takahiro Ogawa, and Miki Haseyama.
\newblock Zero-shot composed image retrieval considering query-target
  relationship leveraging masked image-text pairs.
\newblock \emph{CoRR}, abs/2406.18836, 2024{\natexlab{a}}.
\newblock \doi{10.48550/ARXIV.2406.18836}.
\newblock URL \url{https://doi.org/10.48550/arXiv.2406.18836}.

\bibitem[Radford et~al.(2021)Radford, Kim, Hallacy, Ramesh, Goh, Agarwal,
  Sastry, Askell, Mishkin, Clark, Krueger, and Sutskever]{clip}
Alec Radford, Jong~Wook Kim, Chris Hallacy, Aditya Ramesh, Gabriel Goh,
  Sandhini Agarwal, Girish Sastry, Amanda Askell, Pamela Mishkin, Jack Clark,
  Gretchen Krueger, and Ilya Sutskever.
\newblock Learning transferable visual models from natural language
  supervision.
\newblock In Marina Meila and Tong Zhang, editors, \emph{Proceedings of the
  38th International Conference on Machine Learning, {ICML} 2021, 18-24 July
  2021, Virtual Event}, volume 139 of \emph{Proceedings of Machine Learning
  Research}, pages 8748--8763. {PMLR}, 2021.
\newblock URL \url{http://proceedings.mlr.press/v139/radford21a.html}.

\bibitem[Li et~al.(2023)Li, Li, Savarese, and Hoi]{li2023blip}
Junnan Li, Dongxu Li, Silvio Savarese, and Steven C.~H. Hoi.
\newblock {BLIP-2:} bootstrapping language-image pre-training with frozen image
  encoders and large language models.
\newblock In Andreas Krause, Emma Brunskill, Kyunghyun Cho, Barbara Engelhardt,
  Sivan Sabato, and Jonathan Scarlett, editors, \emph{International Conference
  on Machine Learning, {ICML} 2023, 23-29 July 2023, Honolulu, Hawaii, {USA}},
  volume 202 of \emph{Proceedings of Machine Learning Research}, pages
  19730--19742. {PMLR}, 2023.
\newblock URL \url{https://proceedings.mlr.press/v202/li23q.html}.

\bibitem[Li et~al.(2021)Li, Selvaraju, Gotmare, Joty, Xiong, and Hoi]{albef}
Junnan Li, Ramprasaath~R. Selvaraju, Akhilesh Gotmare, Shafiq~R. Joty, Caiming
  Xiong, and Steven~Chu{-}Hong Hoi.
\newblock Align before fuse: Vision and language representation learning with
  momentum distillation.
\newblock In Marc'Aurelio Ranzato, Alina Beygelzimer, Yann~N. Dauphin, Percy
  Liang, and Jennifer~Wortman Vaughan, editors, \emph{Advances in Neural
  Information Processing Systems 34: Annual Conference on Neural Information
  Processing Systems 2021, NeurIPS 2021, December 6-14, 2021, virtual}, pages
  9694--9705, 2021.
\newblock URL
  \url{https://proceedings.neurips.cc/paper/2021/hash/505259756244493872b7709a8a01b536-Abstract.html}.

\bibitem[Ji et~al.(2023)Ji, Tu, Jiang, Kong, Cai, Zhao, Wang, Yang, and
  Liu]{scl_vl}
Yatai Ji, Rongcheng Tu, Jie Jiang, Weijie Kong, Chengfei Cai, Wenzhe Zhao,
  Hongfa Wang, Yujiu Yang, and Wei Liu.
\newblock Seeing what you miss: Vision-language pre-training with semantic
  completion learning.
\newblock In \emph{{IEEE/CVF} Conference on Computer Vision and Pattern
  Recognition, {CVPR} 2023, Vancouver, BC, Canada, June 17-24, 2023}, pages
  6789--6798. {IEEE}, 2023.
\newblock \doi{10.1109/CVPR52729.2023.00656}.
\newblock URL \url{https://doi.org/10.1109/CVPR52729.2023.00656}.

\bibitem[Tu et~al.(2023{\natexlab{d}})Tu, Ji, Jiang, Kong, Cai, Zhao, Wang,
  Yang, and Liu]{glscl}
Rong{-}Cheng Tu, Yatai Ji, Jie Jiang, Weijie Kong, Chengfei Cai, Wenzhe Zhao,
  Hongfa Wang, Yujiu Yang, and Wei Liu.
\newblock Global and local semantic completion learning for vision-language
  pre-training.
\newblock \emph{CoRR}, abs/2306.07096, 2023{\natexlab{d}}.
\newblock \doi{10.48550/ARXIV.2306.07096}.
\newblock URL \url{https://doi.org/10.48550/arXiv.2306.07096}.

\bibitem[Hu et~al.(2024)Hu, Tu, Han, He, Cui, Long, Zheng, Fang, Huang, Zhao,
  Zhang, Thai, Zhang, Wang, Yao, Zhao, Zhou, Cai, Zhai, Ding, Jia, Zeng, Li,
  Liu, and Sun]{minicpm_llm}
Shengding Hu, Yuge Tu, Xu~Han, Chaoqun He, Ganqu Cui, Xiang Long, Zhi Zheng,
  Yewei Fang, Yuxiang Huang, Weilin Zhao, Xinrong Zhang, Zheng~Leng Thai,
  Kaihuo Zhang, Chongyi Wang, Yuan Yao, Chenyang Zhao, Jie Zhou, Jie Cai,
  Zhongwu Zhai, Ning Ding, Chao Jia, Guoyang Zeng, Dahai Li, Zhiyuan Liu, and
  Maosong Sun.
\newblock Minicpm: Unveiling the potential of small language models with
  scalable training strategies, 2024.
\newblock URL \url{https://arxiv.org/abs/2404.06395}.

\bibitem[Touvron et~al.(2023)Touvron, Lavril, Izacard, Martinet, Lachaux,
  Lacroix, Rozi{\`{e}}re, Goyal, Hambro, Azhar, Rodriguez, Joulin, Grave, and
  Lample]{llama}
Hugo Touvron, Thibaut Lavril, Gautier Izacard, Xavier Martinet, Marie{-}Anne
  Lachaux, Timoth{\'{e}}e Lacroix, Baptiste Rozi{\`{e}}re, Naman Goyal, Eric
  Hambro, Faisal Azhar, Aur{\'{e}}lien Rodriguez, Armand Joulin, Edouard Grave,
  and Guillaume Lample.
\newblock Llama: Open and efficient foundation language models.
\newblock \emph{CoRR}, abs/2302.13971, 2023.
\newblock \doi{10.48550/ARXIV.2302.13971}.
\newblock URL \url{https://doi.org/10.48550/arXiv.2302.13971}.

\bibitem[Yao et~al.(2024)Yao, Yu, Zhang, Wang, Cui, Zhu, Cai, Li, Zhao, He,
  Chen, Zhou, Zou, Zhang, Hu, Zheng, Zhou, Cai, Han, Zeng, Li, Liu, and
  Sun]{minicpm}
Yuan Yao, Tianyu Yu, Ao~Zhang, Chongyi Wang, Junbo Cui, Hongji Zhu, Tianchi
  Cai, Haoyu Li, Weilin Zhao, Zhihui He, Qianyu Chen, Huarong Zhou, Zhensheng
  Zou, Haoye Zhang, Shengding Hu, Zhi Zheng, Jie Zhou, Jie Cai, Xu~Han, Guoyang
  Zeng, Dahai Li, Zhiyuan Liu, and Maosong Sun.
\newblock Minicpm-v: {A} {GPT-4V} level {MLLM} on your phone.
\newblock \emph{CoRR}, abs/2408.01800, 2024.
\newblock \doi{10.48550/ARXIV.2408.01800}.
\newblock URL \url{https://doi.org/10.48550/arXiv.2408.01800}.

\bibitem[Dong et~al.(2024)Dong, Zhang, Zang, Cao, Wang, Ouyang, Wei, Zhang,
  Duan, Cao, Zhang, Li, Yan, Gao, Zhang, Li, Li, Chen, He, Zhang, Qiao, Lin,
  and Wang]{interlm}
Xiaoyi Dong, Pan Zhang, Yuhang Zang, Yuhang Cao, Bin Wang, Linke Ouyang, Xilin
  Wei, Songyang Zhang, Haodong Duan, Maosong Cao, Wenwei Zhang, Yining Li, Hang
  Yan, Yang Gao, Xinyue Zhang, Wei Li, Jingwen Li, Kai Chen, Conghui He,
  Xingcheng Zhang, Yu~Qiao, Dahua Lin, and Jiaqi Wang.
\newblock Internlm-xcomposer2: Mastering free-form text-image composition and
  comprehension in vision-language large model, 2024.
\newblock URL \url{https://arxiv.org/abs/2401.16420}.

\bibitem[Wang et~al.(2024)Wang, Chen, Wang, Cao, Liu, Gao, Zhu, Zhu, Lu, Qiao,
  and Dai]{wang2024mpo}
Weiyun Wang, Zhe Chen, Wenhai Wang, Yue Cao, Yangzhou Liu, Zhangwei Gao, Jinguo
  Zhu, Xizhou Zhu, Lewei Lu, Yu~Qiao, and Jifeng Dai.
\newblock Enhancing the reasoning ability of multimodal large language models
  via mixed preference optimization.
\newblock \emph{CoRR}, abs/2411.10442, 2024.
\newblock \doi{10.48550/ARXIV.2411.10442}.
\newblock URL \url{https://doi.org/10.48550/arXiv.2411.10442}.

\bibitem[Chen et~al.(2023)Chen, Wu, Wang, Su, Chen, Xing, Zhong, Zhang, Zhu,
  Lu, Li, Luo, Lu, Qiao, and Dai]{internvl}
Zhe Chen, Jiannan Wu, Wenhai Wang, Weijie Su, Guo Chen, Sen Xing, Muyan Zhong,
  Qinglong Zhang, Xizhou Zhu, Lewei Lu, Bin Li, Ping Luo, Tong Lu, Yu~Qiao, and
  Jifeng Dai.
\newblock Internvl: Scaling up vision foundation models and aligning for
  generic visual-linguistic tasks.
\newblock \emph{CoRR}, abs/2312.14238, 2023.
\newblock \doi{10.48550/ARXIV.2312.14238}.
\newblock URL \url{https://doi.org/10.48550/arXiv.2312.14238}.

\bibitem[Tu et~al.(2024)Tu, Sun, Jin, Liao, Huang, and Tao]{spagent}
Rong{-}Cheng Tu, Wenhao Sun, Zhao Jin, Jingyi Liao, Jiaxing Huang, and Dacheng
  Tao.
\newblock Spagent: Adaptive task decomposition and model selection for general
  video generation and editing.
\newblock \emph{CoRR}, abs/2411.18983, 2024.
\newblock \doi{10.48550/ARXIV.2411.18983}.
\newblock URL \url{https://doi.org/10.48550/arXiv.2411.18983}.

\bibitem[Luo et~al.(2025)Luo, Zhang, Yuan, Zhao, Yang, Gu, Wu, Chen, Qiao,
  Long, Tu, Luo, Ju, Xiao, Wang, Xiao, Liu, Yuan, Zhang, Jin, Zhang, Wu, Zhao,
  Tao, Yu, and Zhang]{llma_survey}
Junyu Luo, Weizhi Zhang, Ye~Yuan, Yusheng Zhao, Junwei Yang, Yiyang Gu, Bohan
  Wu, Binqi Chen, Ziyue Qiao, Qingqing Long, Rongcheng Tu, Xiao Luo, Wei Ju,
  Zhiping Xiao, Yifan Wang, Meng Xiao, Chenwu Liu, Jingyang Yuan, Shichang
  Zhang, Yiqiao Jin, Fan Zhang, Xian Wu, Hanqing Zhao, Dacheng Tao, Philip~S.
  Yu, and Ming Zhang.
\newblock Large language model agent: {A} survey on methodology, applications
  and challenges.
\newblock \emph{CoRR}, abs/2503.21460, 2025.
\newblock \doi{10.48550/ARXIV.2503.21460}.
\newblock URL \url{https://doi.org/10.48550/arXiv.2503.21460}.

\bibitem[Ma et~al.(2024)Ma, Lan, Tu, Hu, Huang, and Mao]{m2ramg}
Zi{-}Ao Ma, Tian Lan, Rong{-}Cheng Tu, Yong Hu, Heyan Huang, and Xian{-}Ling
  Mao.
\newblock Multi-modal retrieval augmented multi-modal generation: {A}
  benchmark, evaluate metrics and strong baselines.
\newblock \emph{CoRR}, abs/2411.16365, 2024.
\newblock \doi{10.48550/ARXIV.2411.16365}.
\newblock URL \url{https://doi.org/10.48550/arXiv.2411.16365}.

\bibitem[Li et~al.(2024)Li, Fan, Wong, Yang, and Kankanhalli]{mcl}
Wei Li, Hehe Fan, Yongkang Wong, Yi~Yang, and Mohan~S. Kankanhalli.
\newblock Improving context understanding in multimodal large language models
  via multimodal composition learning.
\newblock In \emph{Forty-first International Conference on Machine Learning,
  {ICML} 2024, Vienna, Austria, July 21-27, 2024}. OpenReview.net, 2024.
\newblock URL \url{https://openreview.net/forum?id=Nm6jYZsBum}.

\bibitem[Koh et~al.(2023)Koh, Salakhutdinov, and Fried]{fromage}
Jing~Yu Koh, Ruslan Salakhutdinov, and Daniel Fried.
\newblock Grounding language models to images for multimodal inputs and
  outputs.
\newblock In Andreas Krause, Emma Brunskill, Kyunghyun Cho, Barbara Engelhardt,
  Sivan Sabato, and Jonathan Scarlett, editors, \emph{International Conference
  on Machine Learning, {ICML} 2023, 23-29 July 2023, Honolulu, Hawaii, {USA}},
  volume 202 of \emph{Proceedings of Machine Learning Research}, pages
  17283--17300. {PMLR}, 2023.
\newblock URL \url{https://proceedings.mlr.press/v202/koh23a.html}.

\bibitem[Zhong et~al.(2024)Zhong, An, Jiang, Ma, Guo, and Huang]{instructcir}
Wenliang Zhong, Weizhi An, Feng Jiang, Hehuan Ma, Yuzhi Guo, and Junzhou Huang.
\newblock Compositional image retrieval via instruction-aware contrastive
  learning.
\newblock \emph{CoRR}, abs/2412.05756, 2024.
\newblock \doi{10.48550/ARXIV.2412.05756}.
\newblock URL \url{https://doi.org/10.48550/arXiv.2412.05756}.

\bibitem[Tu et~al.(2021{\natexlab{a}})Tu, Mao, Guo, Wei, and Huang]{psldh}
Rong{-}Cheng Tu, Xian{-}Ling Mao, Jia{-}Nan Guo, Wei Wei, and Heyan Huang.
\newblock Partial-softmax loss based deep hashing.
\newblock In Jure Leskovec, Marko Grobelnik, Marc Najork, Jie Tang, and Leila
  Zia, editors, \emph{{WWW} '21: The Web Conference 2021, Virtual Event /
  Ljubljana, Slovenia, April 19-23, 2021}, pages 2869--2878. {ACM} / {IW3C2},
  2021{\natexlab{a}}.
\newblock \doi{10.1145/3442381.3449825}.
\newblock URL \url{https://doi.org/10.1145/3442381.3449825}.

\bibitem[Tu et~al.(2021{\natexlab{b}})Tu, Mao, Kong, Shao, Li, Wei, and
  Huang]{wglhh}
Rong{-}Cheng Tu, Xian{-}Ling Mao, Cihang Kong, Zihang Shao, Ze{-}Lin Li, Wei
  Wei, and Heyan Huang.
\newblock Weighted gaussian loss based hamming hashing.
\newblock In Heng~Tao Shen, Yueting Zhuang, John~R. Smith, Yang Yang, Pablo
  C{\'{e}}sar, Florian Metze, and Balakrishnan Prabhakaran, editors, \emph{{MM}
  '21: {ACM} Multimedia Conference, Virtual Event, China, October 20 - 24,
  2021}, pages 3409--3417. {ACM}, 2021{\natexlab{b}}.
\newblock \doi{10.1145/3474085.3475498}.
\newblock URL \url{https://doi.org/10.1145/3474085.3475498}.

\bibitem[Tu et~al.(2023{\natexlab{e}})Tu, Mao, Lin, Ji, Qin, Wei, and
  Huang]{uchstm}
Rong{-}Cheng Tu, Xian{-}Ling Mao, Qinghong Lin, Wenjin Ji, Weize Qin, Wei Wei,
  and Heyan Huang.
\newblock Unsupervised cross-modal hashing via semantic text mining.
\newblock \emph{{IEEE} Trans. Multim.}, 25:\penalty0 8946--8957,
  2023{\natexlab{e}}.
\newblock \doi{10.1109/TMM.2023.3243608}.
\newblock URL \url{https://doi.org/10.1109/TMM.2023.3243608}.

\bibitem[Tu et~al.(2023{\natexlab{f}})Tu, Mao, Ji, Wei, and Huang]{daphcm}
Rong{-}Cheng Tu, Xian{-}Ling Mao, Wenjin Ji, Wei Wei, and Heyan Huang.
\newblock Data-aware proxy hashing for cross-modal retrieval.
\newblock In Hsin{-}Hsi Chen, Wei{-}Jou~(Edward) Duh, Hen{-}Hsen Huang,
  Makoto~P. Kato, Josiane Mothe, and Barbara Poblete, editors,
  \emph{Proceedings of the 46th International {ACM} {SIGIR} Conference on
  Research and Development in Information Retrieval, {SIGIR} 2023, Taipei,
  Taiwan, July 23-27, 2023}, pages 686--696. {ACM}, 2023{\natexlab{f}}.
\newblock \doi{10.1145/3539618.3591660}.
\newblock URL \url{https://doi.org/10.1145/3539618.3591660}.

\bibitem[Zhang et~al.(2024{\natexlab{b}})Zhang, Wei, Pang, Qiu, and
  Zhao]{zhang2023enhance}
Gangjian Zhang, Shikui Wei, Huaxin Pang, Shuang Qiu, and Yao Zhao.
\newblock Enhance composed image retrieval via multi-level collaborative
  localization and semantic activeness perception.
\newblock \emph{{IEEE} Trans. Multim.}, 26:\penalty0 916--928,
  2024{\natexlab{b}}.
\newblock \doi{10.1109/TMM.2023.3273466}.
\newblock URL \url{https://doi.org/10.1109/TMM.2023.3273466}.

\bibitem[Tian et~al.(2022)Tian, Newsam, and Boakye]{TianYuxin23Image}
Yuxin Tian, Shawn~D. Newsam, and Kofi Boakye.
\newblock Image search with text feedback by additive attention compositional
  learning.
\newblock \emph{CoRR}, abs/2203.03809, 2022.
\newblock \doi{10.48550/ARXIV.2203.03809}.
\newblock URL \url{https://doi.org/10.48550/arXiv.2203.03809}.

\bibitem[Gu et~al.(2024)Gu, Chun, Kim, Kang, and Yun]{lincir}
Geonmo Gu, Sanghyuk Chun, Wonjae Kim, Yoohoon Kang, and Sangdoo Yun.
\newblock Language-only efficient training of zero-shot composed image
  retrieval.
\newblock In \emph{{IEEE/CVF} Conference on Computer Vision and Pattern
  Recognition, {CVPR} 2024, Seattle, WA, USA, June 16-22, 2024}, pages
  13225--13234. {IEEE}, 2024.
\newblock \doi{10.1109/CVPR52733.2024.01256}.
\newblock URL \url{https://doi.org/10.1109/CVPR52733.2024.01256}.

\bibitem[Karthik et~al.(2024)Karthik, Roth, Mancini, and Akata]{cirevl}
Shyamgopal Karthik, Karsten Roth, Massimiliano Mancini, and Zeynep Akata.
\newblock Vision-by-language for training-free compositional image retrieval.
\newblock In \emph{The Twelfth International Conference on Learning
  Representations, {ICLR} 2024, Vienna, Austria, May 7-11, 2024}.
  OpenReview.net, 2024.
\newblock URL \url{https://openreview.net/forum?id=EDPxCjXzSb}.

\bibitem[Vo et~al.(2019)Vo, Jiang, Sun, Murphy, Li, Fei{-}Fei, and Hays]{tirg}
Nam Vo, Lu~Jiang, Chen Sun, Kevin Murphy, Li{-}Jia Li, Li~Fei{-}Fei, and James
  Hays.
\newblock Composing text and image for image retrieval - an empirical odyssey.
\newblock In \emph{{IEEE} Conference on Computer Vision and Pattern
  Recognition, {CVPR} 2019, Long Beach, CA, USA, June 16-20, 2019}, pages
  6439--6448. Computer Vision Foundation / {IEEE}, 2019.
\newblock \doi{10.1109/CVPR.2019.00660}.
\newblock URL
  \url{http://openaccess.thecvf.com/content\_CVPR\_2019/html/Vo\_Composing\_Text\_and\_Image\_for\_Image\_Retrieval\_-\_an\_Empirical\_CVPR\_2019\_paper.html}.

\bibitem[Wen et~al.(2023)Wen, Zhang, Song, Wei, and Nie]{tgcir}
Haokun Wen, Xian Zhang, Xuemeng Song, Yinwei Wei, and Liqiang Nie.
\newblock Target-guided composed image retrieval.
\newblock In \emph{Proceedings of the {ACM} International Conference on
  Multimedia}, pages 915--923. {ACM}, 2023.

\bibitem[Wen et~al.(2024{\natexlab{b}})Wen, Song, Chen, Wei, Nie, and
  Chua]{wen2024simple}
Haokun Wen, Xuemeng Song, Xiaolin Chen, Yinwei Wei, Liqiang Nie, and Tat{-}Seng
  Chua.
\newblock Simple but effective raw-data level multimodal fusion for composed
  image retrieval.
\newblock In Grace~Hui Yang, Hongning Wang, Sam Han, Claudia Hauff, Guido
  Zuccon, and Yi~Zhang, editors, \emph{Proceedings of the 47th International
  {ACM} {SIGIR} Conference on Research and Development in Information
  Retrieval, {SIGIR} 2024, Washington DC, USA, July 14-18, 2024}, pages
  229--239. {ACM}, 2024{\natexlab{b}}.
\newblock \doi{10.1145/3626772.3657727}.
\newblock URL \url{https://doi.org/10.1145/3626772.3657727}.

\bibitem[Bai et~al.(2023)Bai, Xu, Liu, Khan, Khan, Zuo, Goh, and
  Feng]{bai2023sentence}
Yang Bai, Xinxing Xu, Yong Liu, Salman Khan, Fahad Khan, Wangmeng Zuo, Rick
  Siow~Mong Goh, and Chun-Mei Feng.
\newblock Sentence-level prompts benefit composed image retrieval.
\newblock \emph{arXiv preprint arXiv:2310.05473}, 2023.

\bibitem[Agnolucci et~al.(2024)Agnolucci, Baldrati, Bertini, and
  Bimbo]{isearle}
Lorenzo Agnolucci, Alberto Baldrati, Marco Bertini, and Alberto~Del Bimbo.
\newblock isearle: Improving textual inversion for zero-shot composed image
  retrieval.
\newblock \emph{CoRR}, abs/2405.02951, 2024.
\newblock \doi{10.48550/ARXIV.2405.02951}.
\newblock URL \url{https://doi.org/10.48550/arXiv.2405.02951}.

\bibitem[Bao et~al.(2025)Bao, Liu, Xu, Zheng, and Xu]{mllmi2w}
Tong Bao, Che Liu, Derong Xu, Zhi Zheng, and Tong Xu.
\newblock {MLLM-I2W:} harnessing multimodal large language model for zero-shot
  composed image retrieval.
\newblock In Owen Rambow, Leo Wanner, Marianna Apidianaki, Hend Al{-}Khalifa,
  Barbara~Di Eugenio, and Steven Schockaert, editors, \emph{Proceedings of the
  31st International Conference on Computational Linguistics, {COLING} 2025,
  Abu Dhabi, UAE, January 19-24, 2025}, pages 1839--1849. Association for
  Computational Linguistics, 2025.
\newblock URL \url{https://aclanthology.org/2025.coling-main.125/}.

\bibitem[Tang et~al.(2024)Tang, Yu, Gai, Zhuang, Xiong, Hu, and Wu]{contex12w}
Yuanmin Tang, Jing Yu, Keke Gai, Jiamin Zhuang, Gang Xiong, Yue Hu, and Qi~Wu.
\newblock Context-i2w: Mapping images to context-dependent words for accurate
  zero-shot composed image retrieval.
\newblock In Michael~J. Wooldridge, Jennifer~G. Dy, and Sriraam Natarajan,
  editors, \emph{Thirty-Eighth {AAAI} Conference on Artificial Intelligence,
  {AAAI} 2024, Thirty-Sixth Conference on Innovative Applications of Artificial
  Intelligence, {IAAI} 2024, Fourteenth Symposium on Educational Advances in
  Artificial Intelligence, {EAAI} 2014, February 20-27, 2024, Vancouver,
  Canada}, pages 5180--5188. {AAAI} Press, 2024.
\newblock \doi{10.1609/AAAI.V38I6.28324}.
\newblock URL \url{https://doi.org/10.1609/aaai.v38i6.28324}.

\bibitem[Sharma et~al.(2018)Sharma, Ding, Goodman, and Soricut]{cc3m}
Piyush Sharma, Nan Ding, Sebastian Goodman, and Radu Soricut.
\newblock Conceptual captions: {A} cleaned, hypernymed, image alt-text dataset
  for automatic image captioning.
\newblock In Iryna Gurevych and Yusuke Miyao, editors, \emph{Proceedings of the
  56th Annual Meeting of the Association for Computational Linguistics, {ACL}
  2018, Melbourne, Australia, July 15-20, 2018, Volume 1: Long Papers}, pages
  2556--2565. Association for Computational Linguistics, 2018.
\newblock \doi{10.18653/V1/P18-1238}.
\newblock URL \url{https://aclanthology.org/P18-1238/}.

\bibitem[van~den Oord et~al.(2018)van~den Oord, Li, and Vinyals]{infonce}
A{\"{a}}ron van~den Oord, Yazhe Li, and Oriol Vinyals.
\newblock Representation learning with contrastive predictive coding.
\newblock \emph{CoRR}, abs/1807.03748, 2018.
\newblock URL \url{http://arxiv.org/abs/1807.03748}.

\bibitem[Wu et~al.(2021)Wu, Gao, Guo, Al{-}Halah, Rennie, Grauman, and
  Feris]{fashioniq}
Hui Wu, Yupeng Gao, Xiaoxiao Guo, Ziad Al{-}Halah, Steven Rennie, Kristen
  Grauman, and Rog{\'{e}}rio Feris.
\newblock Fashion {IQ:} {A} new dataset towards retrieving images by natural
  language feedback.
\newblock In \emph{{IEEE} Conference on Computer Vision and Pattern
  Recognition, {CVPR} 2021, virtual, June 19-25, 2021}, pages 11307--11317.
  Computer Vision Foundation / {IEEE}, 2021.
\newblock \doi{10.1109/CVPR46437.2021.01115}.
\newblock URL
  \url{https://openaccess.thecvf.com/content/CVPR2021/html/Wu\_Fashion\_IQ\_A\_New\_Dataset\_Towards\_Retrieving\_Images\_by\_Natural\_CVPR\_2021\_paper.html}.

\bibitem[Russakovsky et~al.(2015)Russakovsky, Deng, Su, Krause, Satheesh, Ma,
  Huang, Karpathy, Khosla, Bernstein, Berg, and Fei{-}Fei]{imagenet}
Olga Russakovsky, Jia Deng, Hao Su, Jonathan Krause, Sanjeev Satheesh, Sean Ma,
  Zhiheng Huang, Andrej Karpathy, Aditya Khosla, Michael~S. Bernstein,
  Alexander~C. Berg, and Li~Fei{-}Fei.
\newblock Imagenet large scale visual recognition challenge.
\newblock \emph{Int. J. Comput. Vis.}, 115\penalty0 (3):\penalty0 211--252,
  2015.
\newblock \doi{10.1007/S11263-015-0816-Y}.
\newblock URL \url{https://doi.org/10.1007/s11263-015-0816-y}.

\bibitem[Loshchilov and Hutter(2019)]{adamw}
Ilya Loshchilov and Frank Hutter.
\newblock Decoupled weight decay regularization.
\newblock In \emph{7th International Conference on Learning Representations,
  {ICLR} 2019, New Orleans, LA, USA, May 6-9, 2019}. OpenReview.net, 2019.
\newblock URL \url{https://openreview.net/forum?id=Bkg6RiCqY7}.

\bibitem[Lin et~al.(2014)Lin, Maire, Belongie, Hays, Perona, Ramanan,
  Doll{\'{a}}r, and Zitnick]{coco}
Tsung{-}Yi Lin, Michael Maire, Serge~J. Belongie, James Hays, Pietro Perona,
  Deva Ramanan, Piotr Doll{\'{a}}r, and C.~Lawrence Zitnick.
\newblock Microsoft {COCO:} common objects in context.
\newblock In David~J. Fleet, Tom{\'{a}}s Pajdla, Bernt Schiele, and Tinne
  Tuytelaars, editors, \emph{Computer Vision - {ECCV} 2014 - 13th European
  Conference, Zurich, Switzerland, September 6-12, 2014, Proceedings, Part
  {V}}, volume 8693 of \emph{Lecture Notes in Computer Science}, pages
  740--755. Springer, 2014.
\newblock \doi{10.1007/978-3-319-10602-1\_48}.
\newblock URL \url{https://doi.org/10.1007/978-3-319-10602-1\_48}.

\bibitem[Suhr et~al.(2019)Suhr, Zhou, Zhang, Zhang, Bai, and Artzi]{nlvr2}
Alane Suhr, Stephanie Zhou, Ally Zhang, Iris Zhang, Huajun Bai, and Yoav Artzi.
\newblock A corpus for reasoning about natural language grounded in
  photographs.
\newblock In Anna Korhonen, David~R. Traum, and Llu{\'{\i}}s M{\`{a}}rquez,
  editors, \emph{Proceedings of the 57th Conference of the Association for
  Computational Linguistics, {ACL} 2019, Florence, Italy, July 28- August 2,
  2019, Volume 1: Long Papers}, pages 6418--6428. Association for Computational
  Linguistics, 2019.
\newblock \doi{10.18653/V1/P19-1644}.
\newblock URL \url{https://doi.org/10.18653/v1/p19-1644}.

\end{thebibliography}

\appendix


\newpage

\section{Limitations}
While MLLMs exhibit strong visual understanding and semantic reasoning capabilities, they are still prone to hallucinations and other generation errors. Consequently, when using an MLLM to construct training triplets and pairs for fine-tuning the VLM, some incorrect data associations may be introduced, leading to noise that can hinder retrieval performance. In future work, we will explore strategies to mitigate the impact of noisy data pairs and enhance the reliability of generated supervision signals.

\section{Broader Impact}
This work proposes a scalable, annotation-free framework for zero-shot composed image retrieval (ZS-CIR), which may benefit domains such as e-commerce, content creation, and education by enabling intuitive image search via visual-textual modifications. Leveraging MLLM-generated supervision helps reduce annotation cost and increases accessibility in low-resource settings.

However, our method also presents potential risks. Enhanced retrieval models may be misused in surveillance or misinformation workflows. Moreover, reliance on pretrained MLLMs may propagate societal biases present in the synthetic supervision. To mitigate these issues, we use only publicly available datasets, acknowledge model limitations, and encourage future applications to incorporate transparency tools, bias auditing, and uncertainty estimation mechanisms.

\section{Evaluation datasets and BLIP2 based Baseline Variants}
\label{detial}
In our experiments, we adopt three widely used public CIR benchmarks: \textbf{FashionIQ} \cite{fashioniq}, \textbf{CIRCO} \cite{searle}, and \textbf{CIRR} \cite{cirr}. Each dataset emphasizes different aspects of composed image retrieval:
\begin{itemize}[nosep]
\item \textbf{FashionIQ.}
This dataset focuses on fashion items across three categories: Dress, Shirt, and Tops\&Tee. It provides 36k validation triplets but no public test set. Following prior work \cite{fti4cir,searle,contex12w}, we evaluate on the validation set and report retrieval performance over the three categories.
\item \textbf{CIRCO.}
Built upon the COCO dataset \cite{coco}, CIRCO introduces multiple ground-truth targets per sample to reflect the fact that more than one image can satisfy a given modification text. Each triplet consists of a reference image, a piece of modifying text, and multiple valid target images. We use the official test set with 800 samples to assess multi-target retrieval scenarios, which underscores the ability of a CIR model to handle non-unique solutions.
\item \textbf{CIRR.}
Based on the NLVR2 dataset \cite{nlvr2}, CIRR contains approximately 21k real-world images, each annotated with a modifying text that uniquely corresponds to one target image. This design reduces the risk of false negatives, but it also imposes more stringent requirements on the model’s compositional reasoning. We adopt the official CIRR test set (4.1k triplets) for evaluation.
\end{itemize}
The four baseline variants using BLIP2 encoders are defined as follows :
\begin{itemize}[nosep]
\item \textbf{Image-only}: Encode both the reference image and a candidate image using BLIP2’s visual encoder and directly compute their feature similarity;
\item \textbf{Text-only}: Encode the modifying text and the candidate image using BLIP2, then measure their similarity; 
\item \textbf{Image + Text}: Compute a unified query representation by averaging the features of the reference image and modifying text, then evaluate its similarity with the candidate image features; 
\item \textbf{Target Text}: Generate an imagined target text using the MLLM via Eq.~(\ref{inference_tt}), and encode it with BLIP2’s text encoder, and then compute its similarity with the candidate images.
\end{itemize}

\section{Data Curation Prompt Templates}
\label{sec:prompt}
In here, we illustrate the prompts for guiding the MLLM to generate modification text, target text and image caption in the Figure \ref{fig:mofication_generate},  \ref{fig:target_text} and  \ref{fig:caption}, respectively. Additionally, the corresponding prompts $\boldsymbol{P}'$ are shown in Figure \ref{fig:mofication_generate1},  \ref{fig:target_text1} and  \ref{fig:caption1}, respectively.

\begin{figure}[]
\scriptsize
\centering
\begin{tcolorbox}
\textbf{\# Task Description}\\
You are an expert in understanding and modifying images of objects. Your task is to generate one concise modification text for the given image based on its content. The modification should focus specifically on the object's attributes, including but not limited to:
\begin{itemize}[nosep]
    \renewcommand{\labelitemi}{-}
    \item Change the object's color, material, texture, or pattern.
    \item Adjust the object's shape, size, or structural design.
    \item Modify specific features of the object, such as handles, edges, or attachments.
    \item Add or remove design elements like patterns, decorations, or markings.
    \item Transform the object’s overall appearance (e.g., from modern to antique, or from sleek to rugged).
    \item Combine any of the above changes or introduce other creative adjustments to the object. \\
\end{itemize}

Provide the modification text in one clear and concise sentence without any explanation or additional context.\\
\end{tcolorbox}
\caption{Prompt template $P_m$ for modification text generation.}
\label{fig:mofication_generate}
\end{figure}

\begin{figure}[]
\scriptsize
\centering
\begin{tcolorbox}
\textbf{\# Task Description}\\
You are an expert in object image editing and visualization. Your task is to imagine how the object in the given image would look after being modified according to the description "\{modification\_text\}". Write exactly one clear and concise sentence describing only the modified image, focusing on the most important object details, such as:
\begin{itemize}[nosep]
    \renewcommand{\labelitemi}{-}
    \item The type and category of the object.
    \item The color, material, texture, and pattern of the object.
    \item Distinctive design features, shapes, and structural details.
    \item Any notable attributes that define the object's appearance. \\
\end{itemize}

Provide the description in one clear and complete sentence without referencing the original object, the modification process, or any comparisons.\\
\end{tcolorbox}
\caption{Prompt template $P_{tt}$ for target text generation.}
\label{fig:target_text}
\end{figure}

\begin{figure}[]
\scriptsize
\centering
\begin{tcolorbox}
\textbf{\# Task Description}\\
You are an expert in image analysis and description. Your job is to generate one precise and concise sentence that fully describes the content of the given image. Focus on the most important details, such as:
\begin{itemize}[nosep]
    \renewcommand{\labelitemi}{-}
    \item The primary objects or elements in the image.
    \item The relationships, positions, or actions of these objects.
    \item The overall setting, background, or scene type. \\
\end{itemize}

Provide the modification text in one clear and concise sentence without any explanation or additional context.\\
\end{tcolorbox}
\caption{Prompt template $P_c$  for image caption generation.}
\label{fig:caption}
\end{figure}

\begin{figure}[]
\scriptsize
\centering
\begin{tcolorbox}
\textbf{\# Task Description}\\
You are an expert in understanding and modifying image content. Your job is to generate one concise modification text for the given image based on its content. The modification should focus on one or more of the following aspects:
\begin{itemize}[nosep]
    \renewcommand{\labelitemi}{-}
    \item Replace or change the background (e.g., season, environment, or weather).
    \item Change the color, shape, size, quantity, or texture of objects.
    \item Adjust the position, angle, or arrangement of objects.
    \item Add new objects, details, or elements to the scene.
    \item Remove specific objects or parts of the background.
    \item Combine any of the above changes or introduce other creative adjustments.\\
\end{itemize}

Provide the modification text in one clear and concise sentence without any explanation or additional context.\\
\end{tcolorbox}
\caption{Prompt template $P'_m$ for modification text generation in the ablation study.}
\label{fig:mofication_generate1}
\end{figure}

\begin{figure}[]
\scriptsize
\centering
\begin{tcolorbox}
\textbf{\# Task Description}\\
You are an expert in image editing and visualization. Your task is to imagine the content of the image after it has been modified based on the description "{modification\_text}". Write exactly one clear and concise sentence describing the modified image as if it were a new and independent image, focusing on the most important details, such as:
\begin{itemize}[nosep]
    \renewcommand{\labelitemi}{-}
    \item The primary objects or elements in the modification image.
    \item The relationships, positions, or actions of these objects.
    \item The overall setting, background, or scene type. \\
\end{itemize}

Provide the description in one clear and complete sentence without referencing the original image, the modification process, or any comparisons.\\
\end{tcolorbox}
\caption{Prompt template $P'_{tt}$ for target text generation in the ablation study.}
\label{fig:target_text1}
\end{figure}

\begin{figure}[]
\scriptsize
\centering
\begin{tcolorbox}
\textbf{\# Task Description}\\
You are an expert in image analysis and description. Your job is to generate one precise and concise sentence that fully describes the content of the given image. Focus on the most important details, such as:
\begin{itemize}[nosep]
    \renewcommand{\labelitemi}{-}
    \item The primary objects or elements in the image.
    \item The relationships, positions, or actions of these objects.
    \item The overall setting, background, or scene type. \\
\end{itemize}

Provide the modification text in one clear and concise sentence without any explanation or additional context.\\
\end{tcolorbox}
\caption{Prompt template $P'_c$  for image caption generation in the ablation study.}
\label{fig:caption1}
\end{figure}

\begin{figure}[]
    \centering
    \includegraphics[width=\linewidth]{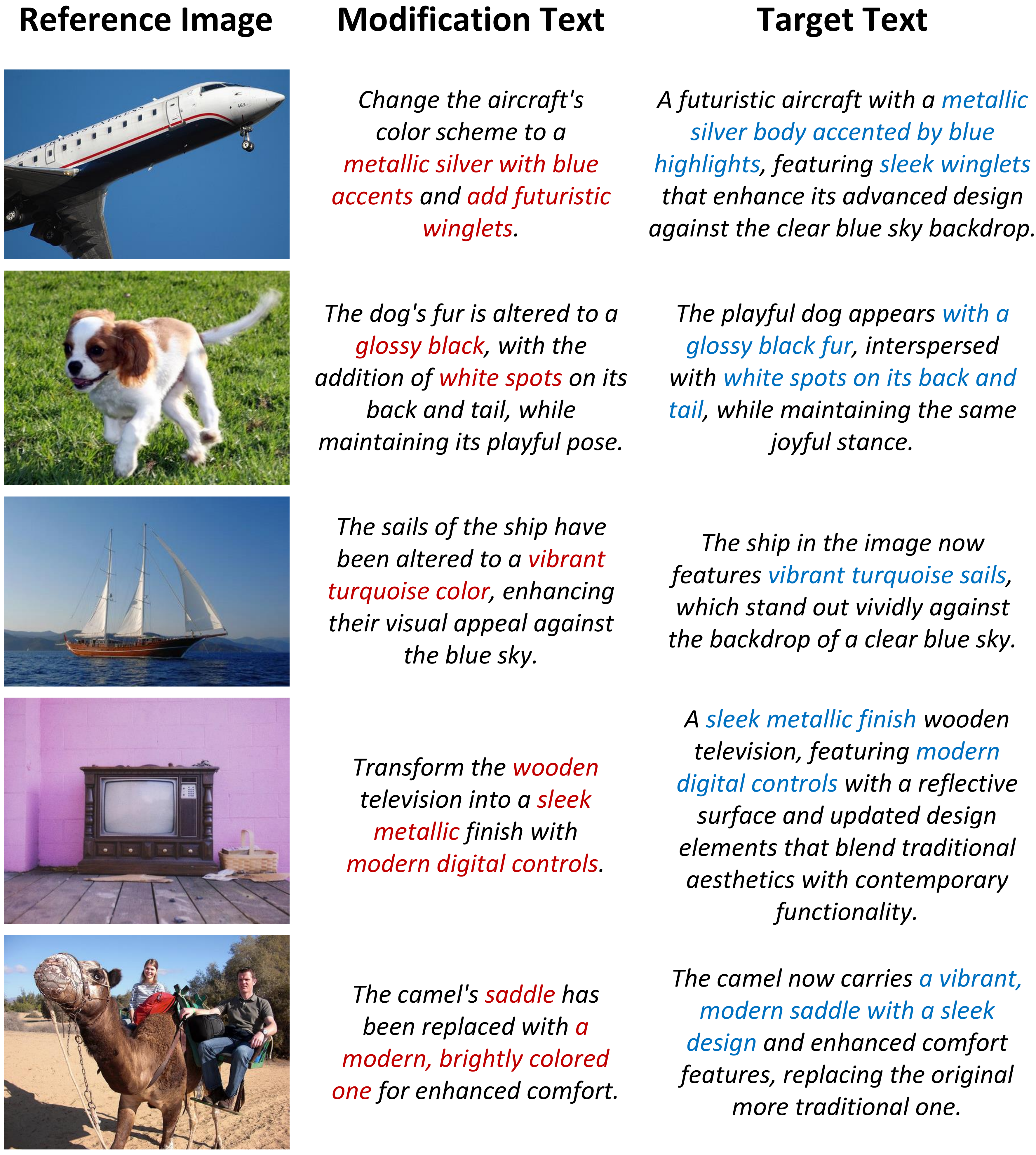}
    \caption{Examples of <Reference Image, Modification Text, Target Text> triplets generated by the MLLM.}
    \label{fig:triplets}
\end{figure}

Additionally, we show some examples of the <\text{Reference Image}, \text{Modification Text}, \text{Target Text}> triplets generated by the multimodal large language model (MLLM) in Figure~\ref{fig:triplets}. Each row depicts a given reference image, a generated modifying text, and a target text automatically produced by the MLLM. Notably, the target text goes beyond merely incorporating the requested modifications: it also maintains contextual details relevant to the original scene. For instance, when asked to ``change the aircraft's color scheme to a metallic silver with blue accents and add futuristic winglets,'' the generated target text not only includes the color alterations but also highlights sleek winglets in a clear sky backdrop. Similarly, in the second example, the dog’s fur is changed to a glossy black with white spots, while the description preserves the dog’s playful posture.

\begin{figure}[]
    \centering
    \includegraphics[width=\linewidth]{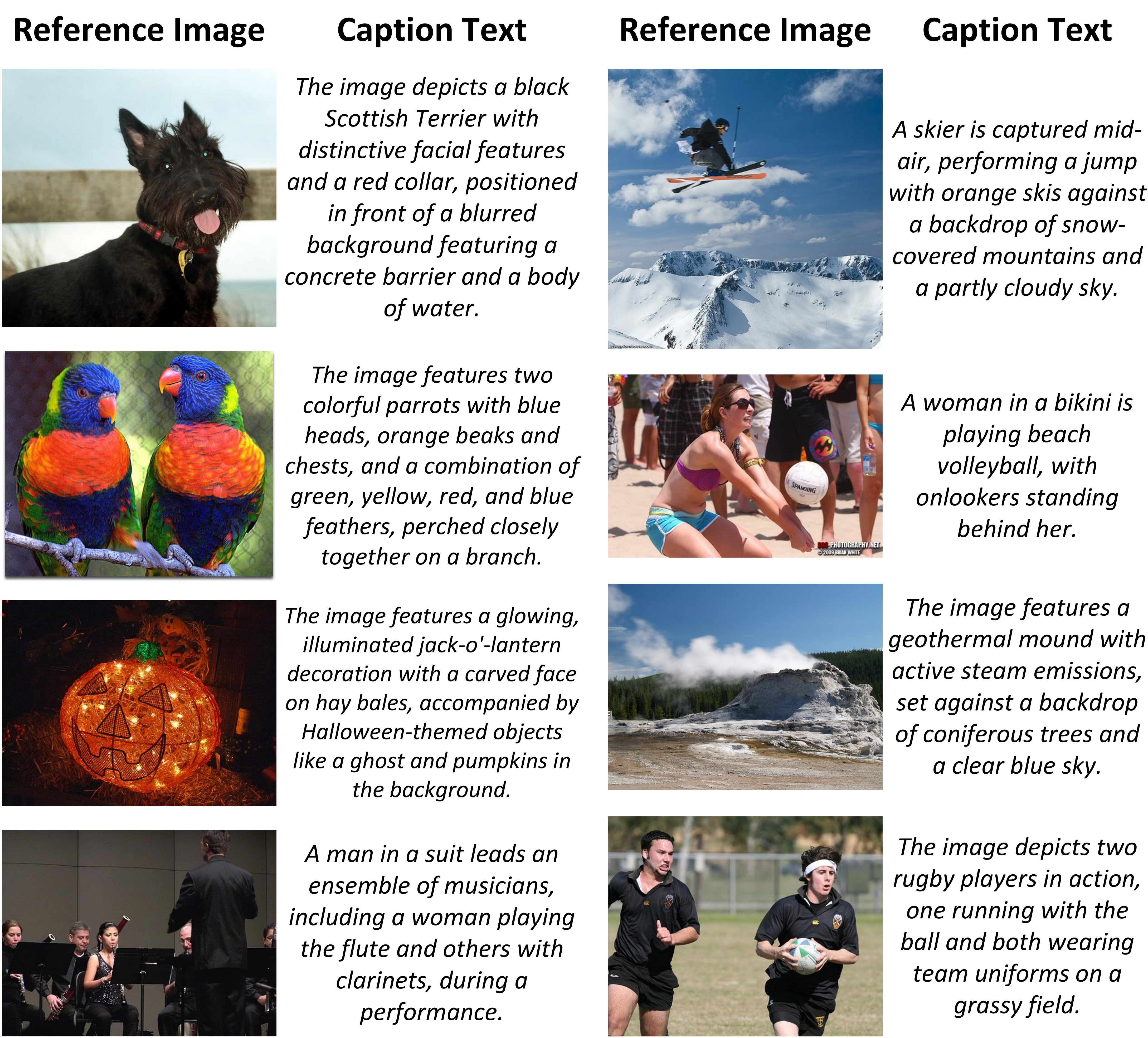}
    \caption{Examples of <Reference Image, Caption Text> generated by the MLLM.}
    \label{fig:image-caption}
\end{figure}

In addition to generating triplets, the MLLM is also emploied to produce image--caption pairs for the unlabeled training images. As illustrated in Figure~\ref{fig:image-caption}, each reference image is accompanied by a contextually rich caption that describes salient visual elements, object attributes, and the overall scene. For instance, the top-left example features a black Scottish Terrier with distinctive facial features against a background of water.

When considered alongside the triplets in Figure~\ref{fig:triplets}, these image--caption pairs serve a complementary role. The triplets focus on how visual content evolves after text-guided modifications, whereas the caption pairs depict how the MLLM captures intrinsic details of each scene without any explicit modification instructions. Together, they form a more comprehensive training resource for zero-shot CIR: the triplets guide the vision-language model (VLM) to learn compositional transformations, and the captions reinforce its ability to associate natural language descriptions with unaltered visual content. This dual strategy maximizes data utility from unlabeled images, enabling our method to align the features of composed queries with their corresponding target images.

\section{Case Study}
\begin{figure*}[]
    \centering
    \includegraphics[width=0.95\linewidth]{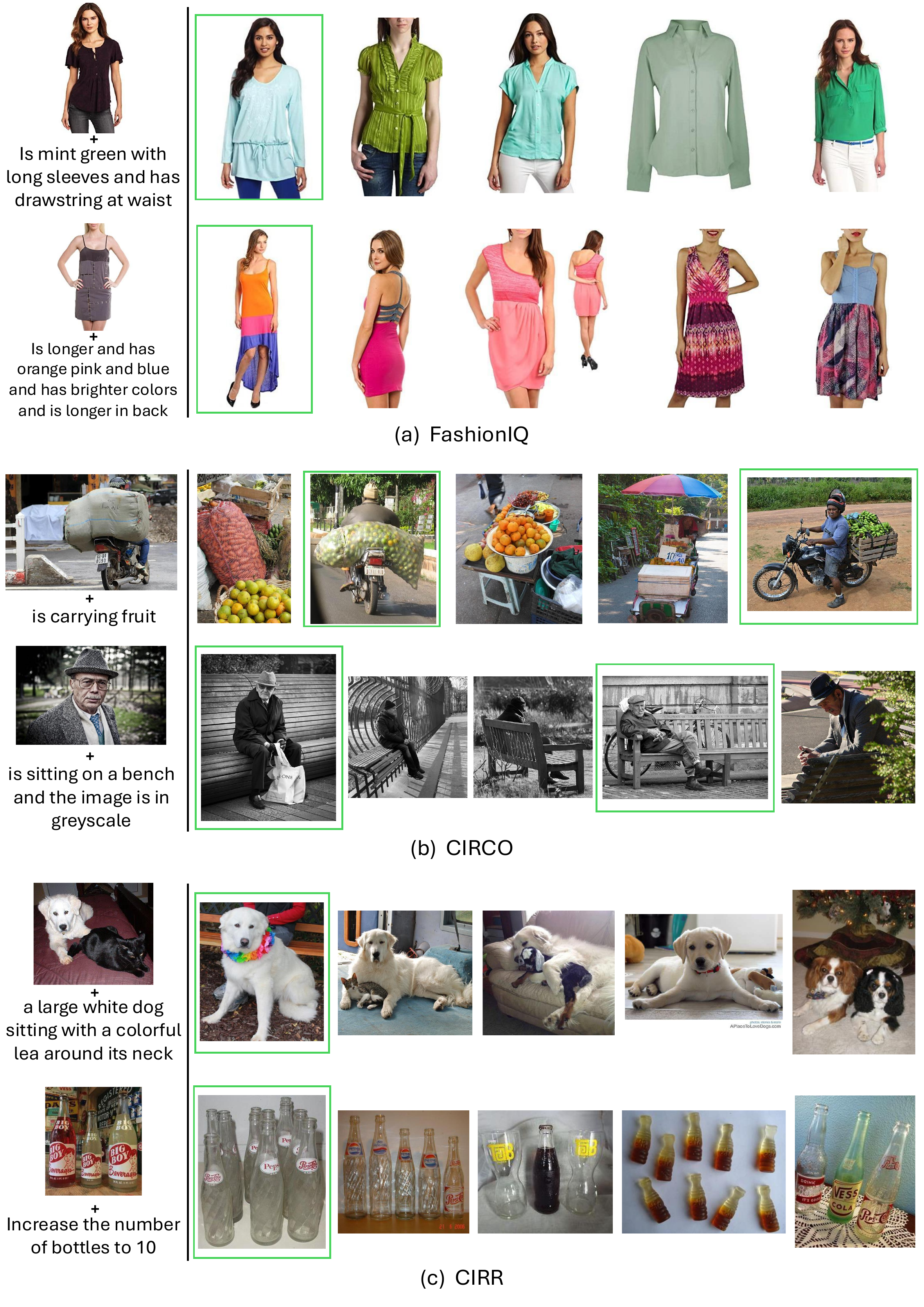}
    \caption{Retrieved results on the three datasets. The green box indicates the target image.}
    \label{fig:addtional_cases}
\end{figure*}

In Figure~\ref{fig:addtional_cases}, we present additional composed image retrieval examples across the three benchmark datasets: FashionIQ, CIRR, and CIRCO. Our method consistently retrieves the correct target images with high accuracy, further demonstrating its effectiveness in Zero-Shot Composed Image Retrieval.

On the FashionIQ and CIRR datasets, our fine-tuned model ranks the correct target image in the top position, showcasing that our approach effectively captures fine-grained compositional semantics, leading to highly precise retrieval results.

For the CIRCO dataset, where each composed query may correspond to multiple valid target images, our model successfully ranks these target images among the top results. As shown in Figure~\ref{fig:addtional_cases}~(b), the retrieved results place the two correct target images in the top five returned results. This demonstrates that our method is capable of handling scenarios where multiple visually relevant targets exist, ranking them higher in the candidate list despite the increased retrieval complexity.

Overall, these qualitative results validate that our approach not only excels at retrieving a single precise target but also generalizes well to cases with multiple correct matches, further highlighting its robustness across diverse datasets.

\end{document}